\documentclass[%
 reprint,
 amsmath,amssymb,
 aps,
]{revtex4-1}

\usepackage{graphicx}
\usepackage{dcolumn}
\usepackage{bm}

\usepackage{algorithm}
\usepackage{algorithmic}

\begin{document}

\preprint{APS/123-QED}

\title{Extracting Dynamical Models from Data}

\author{Michael F. Zimmer}
 \homepage{http://www.neomath.com}

\date{\today}

\begin{abstract}
The problem of determining the underlying dynamics of a system when only given data of its state over time has challenged scientists for decades.  
In this paper, the approach of using machine learning to model the {\em updates} of the phase space variables is introduced; this is done as a function of the phase space variables.  (More generally, the modeling is done over functions of the jet space.)  
This approach (named ``FJet") allows one to accurately replicate the dynamics, and is demonstrated on the examples of the damped harmonic oscillator, the damped pendulum, and the Duffing oscillator;
the underlying differential equation is also accurately recovered for each example.
In addition, the results in no way depend on how the data is sampled over time (i.e., regularly or irregularly).
It is demonstrated that a regression implementation of FJet is similar to the model resulting from a Taylor series expansion of the Runge-Kutta (RK) numerical integration scheme.  
This identification confers the advantage of explicitly revealing the function space to use in the modeling, 
as well as the associated uncertainty quantification for the updates.  
Finally, it is shown in the undamped harmonic oscillator example that the stability of the updates is stable $10^9$ times longer than with $4$th-order RK (with time step $0.1$).
\end{abstract}

\maketitle



\section{Introduction}

Equations of motion for dynamical systems have traditionally been derived by a variety of techniques of a theoretical, experimental, and/or intuitive nature.
However, in the case that differential equations (DEs) are determined directly from a {\em data} source, perhaps with machine learning (ML) techniques, it is useful to identify the following subgoals:
\begin{enumerate}
\item Extrapolate model beyond training data times.
\item Determine underlying DE.
\item Determine parameter dependencies in DE.
\item Estimate stability/accuracy of model.
\item Determine related domain knowledge. 
\end{enumerate}
The present paper focuses on the first four points.  
The fifth point relates to conserved quantities and is discussed in detail in a companion paper \citep{Zimmer-MLDE-arxiv}.

\subsection*{Background}
\label{sec:background}

Related developments in the field appear over the history of the subject, and are organized accordingly.
For completeness, earlier works are included.

{\em \underline{Classical} }:
Systematically deriving DEs for dynamical systems can be identified as beginning with Newton's second law of motion \cite{Goldstein-book-2nd}.
Lagrange recast Newtonian mechanics, building on work of others to take into account constraints.
Lagrangian mechanics are built in terms of the generalized coordinates and velocities ($q$,$\dot{q}$), leading to the Euler-Lagrange equations.
Hamilton gave the modern version of the principle of least action, and after introducing a generalized momentum ($p$),
and a Legendre transform (from $(q,\dot{q}) \rightarrow (q,p)$), produced the Hamiltonian mechanics.
Non-conservative features may be accounted for by including Rayleigh dissipation,
or by a generalization of Hamilton's variational principle \citep{Galley-2013}.

{\em \underline{Nonlinear Dynamics} }:
The beginning of the study of nonlinear dynamics is widely identified with Poincar\'{e}'s work \cite{Poincare-Goff-v1-1993,Holmes-1990}, owing to his studies on celestial mechanics, three-body problems, qualitative studies of nonlinear DEs \cite{Poincare-1881-DEs}, and a wealth of contributions for analyzing such systems (e.g., phase space, recurrence theorem, Poincar\'e-Bendixson theorem).  
Also, Lyapunov \cite{Lyapunov-book-reprint} studied the stability of ordinary DEs (ODEs), introducing an exponential characterization of the rate of growth of instabilities.


{\em \underline{Chaos} }:
Certain nonlinear systems may exhibit a sensitive dependence on initial conditions \cite{Atlee-book-v1, Sprott-book}, so that the long time difference between two nearby states may become arbitrarily large (within the confines of the attractor).
Such a phenomena was first studied by Lorenz \cite{Lorenz-1963} in the context of equations that modeled atmospheric convection.  
A general definition of chaos may be found in \cite[Sec.~1.8 of ][]{Devaney-book}.

The sensitive dependence on initial conditions can be measured by a Lyapunov exponent \citep{Atlee-book-v1,Atlee-book-v2}, 
with a positive value signifying chaos \citep{Geist-1990}.
In numerical simulations, there are thus two sources of instability: from discretization effects, and also from chaos \citep{Crutchfield-1987}.  
The question of whether a solution of a chaotic system is even meaningful was answered in the affirmative with the {\em shadowing lemma} \citep{Bowen-1978,Guckenheimer-book,Palmore-1986,Palmore-1987,Grebogi-1990}.
The length of time that a solution remains valid despite such instabilities was studied by \cite{Sauer-1997}.

{\em \underline{Embeddings} }:
In the context of fluid turbulence, there was interest in testing the Ruelle-Takens conjecture \citep{Ruelle-1971} concerning the role of a strange attractor.
The challenge was to connect experimental time series data at a point to a strange attractor.  
It was shown by Packard et al. \cite{Packard-1980-geometry} that geometric aspects (i.e., Lyapunov exponent, topological dimension) of a R\"ossler model could be captured by using a lower dimensional manifold \citep{Takens-1981} and time-delay coordinates of a single scalar variable $s(t)$ (e.g., density or velocity component) 
at multiple times (cf. Sec 8.13 of \citep{Atlee-book-v2}).
Important related works include \cite{Farmer-1987,Roux-1980,Brandstater-1983}.

{\em \underline{DE Models} }:
A methodology for modeling the global dynamics was introduced by Crutchfield and McNamara \cite{Crutchfield-1987}.
The data was fit to a model of the form $dy/dt = F$, where the choice of $F$ was guided by minimizing a complexity measure.
They considered low-dimensional models, including the Lorenz model and the Duffing oscillator.
A number of other significant works would soon follow, including \citep{Broomhead-1988,Casdagli-1989,Cremers-Hubler-1987,Breeden-1990}.
In particular, \cite{Gouesbet-1991,Gouesbet-1992} would model the dynamics in the phase space, although in a manner different from the present work.
The techniques that were used in this time period for modeling $F(y)$ included polynomial expansions, radial basis functions, neural nets (NNs), B-splines, Galerkin bases \cite{Broomhead-1991}, etc;
the reader is referred to the review by \cite{Aguirre-review-2009} for a more complete discussion.

{\em \underline{Recent} }:
There has been renewed attention to this problem, exploiting techniques from the ML community.
A common theme has been to apply the time series data (from conserved dynamics) to an autoencoder, and to require that the latent variables (i.e., the middle layer) obey physical equations of motion, such as Hamilton's equations \citep{Greydanus-HNN-2019,Bondesan-2019,Toth-2020} or the Euler-Lagrange equations \citep{Lutter-DeLaN-2019,Cranmer-LNN-2020}.

Another approach has been the application of a parametrized symbolic regression approach \citep{Cranmer-2020}, applied to Graph NNs on a cosmology example.

Software (SINDy) \citep{Rudy-2017} has been used to search a function space while modeling, beginning with a regularized form of $dy/dt = F$;
this was demonstrated on a number of ODEs and partial DEs (PDEs).

In addition, Neural ODEs \cite{Chen-2018,Ott-2020,Kidger-2022} can replicate the dynamics of a system by first training a NN on data, and then using that NN for a new extrapolation.
In this approach, each layer of the NN represents an update step in a numerical integration scheme for a DE.

Finally, there are other contributions besides these, but space does not permit a more complete review.

{\em \underline{Other} } Related topics include time-series analysis, numerical integration, and Lie symmetries.

{\em Time-Series Analysis}:
There is a long literature on time-series analysis \citep{BoxJenkins-5th}, and there is significant overlap with what is being focused on here.  
The difference is that with a typical time-series analysis, there is little additional interest in finding the underlying dynamical equations that lead to the data.  
Instead, the focus is often just on finding a statistical characterization or perhaps on fitting a simple model.  
For this reason, no further discussion will be made of this.

{\em Numerical Integration}:
At first glance it would seem that the topic of numerical integration \citep{Hairer-1987,numrec,Schober-2014} wouldn't belong in this section.  However, an expansion with respect to the time-step of a scheme such as 4th-order Runge-Kutta (RK4) produces terms that can reasonably be considered in a model.  Thus it offers a principled means of determining the function space for the predictors of a model, and will be used throughout this paper.
(Of course, the RK scheme is itself based on a Taylor series expansion of a solution (cf. Ch.2 of \citep{Hairer-1987}, or App.B of \citep{Ince-1956}).  
However, the author finds it convenient to make comparisons
to RK directly, since RK is used in other contexts in this paper.)

{\em Lie Symmetries}:
The present work will study the model with respect to small changes in the phase space variables, and sometimes higher derivatives.  
At the same time though, it is adjacent to the effort of determining conserved quantities, such as energy or momentum.  
Lie symmetries were introduced to solve ODEs and PDEs by studying their invariances \citep{Olver-1993}, and will be considered in a companion paper for using the models computed herein to determine conserved quantities.
Interestingly, this can also be done for {\em non-conserved} models.

\subsection*{Outline}

The layout of the remainder of the paper is structured as follows.
Section~\ref{sec:FJet} introduces the FJet approach, going into detail in the Methodology subsection.
In Sections~\ref{sec:HO},  \ref{sec:Pendulum} and \ref{sec:Duffing} are the three examples: Harmonic Oscillator, Pendulum, and Duffing Oscillator.
The analysis in each of these sections follows a similar pattern: training data, model, underlying DE, extrapolation, etc.
Section~\ref{sec:fcnspace} takes a larger view of how Runge-Kutta (RK) informs the choice of the function space to be used for the predictors.
The appearance of an implicit regularization is discussed in Sec.~\ref{sec:regu}.
Following that are the Final Remarks and several appendices.
Appendix~\ref{sec:app-RK} analyzes the RK updates for each model, and shows how the function space of the predictor variables appear after a Taylor series expansion;
these results play an important role in this paper.
Appendix~\ref{sec:app-params} collects results from the parameter fitting, 
and Appendix~\ref{sec:app-errors} displays plots of the residuals.
Finally, Appendix~\ref{sec:app-opt} discusses a second optimization procedure which can improve the quality of a model

\section{FJet Approach}
\label{sec:FJet}

The focus in this paper is on dynamical systems describable as an explicit $n$th-order ODE
\begin{align}
u^{(n)} & = G(t, u, u^{(1)}, ..., u^{(n-1)} )
\label{eqn:scopeeqn}
\end{align}
where $u^{(i)}$ represents the $i$th derivative of $u$ with respect to time $t$.
Of course, neither the order of the equation nor the form of $G$ will be known beforehand.
Both autonomous and non-autonomous cases will be considered.

As mentioned, the approach of related works has been to model $u$ and $u^{(1)}$ as a function of $t$.
Here the approach will be to model {\em changes} in $U^{(n-1)} = \{ u, u^{(1)}, \dots, u^{(n-1)} \}$, known as the $(n-1)$st prolongation \citep{Olver-1993} of $u(t)$.
Specifically, for an $n$-dimensional system (cf. Eq.~\ref{eqn:scopeeqn}), 
the response variables used here are $Y = \Delta {\cal U}^{(n-1)} \equiv \{ \Delta u, \Delta u^{(1)}, \dots, \Delta u^{(n-1)} \}$.
For {\em autonomous} dynamics, the predictor variables are $X = {\cal U}^{(n-1)}$.
For non-autonomous dynamics, the predictors are $X = Q \times {\cal U}^{(n-1)}$, otherwise known as the $(n-1)$st jet space, where $Q$ is the set of independent variables (in this paper just $\{ t \}$).
Also, in one example of this paper where a function of $Q$ appears, the role of $Q$ is supplanted by a prolongation of that function.
Specifically, in Sec.~\ref{sec:Duffing} rather than using $X = \{ t \} \times {\cal U}^{(n-1)}$ for the predictor variables, the space $X = \{ p, p^{(1)}, \dots \} \times {\cal U}^{(n-1)}$ is used, where $p = p(t)$.
Finally, for both cases (autonomous and non-autonomous), functions of the elements of $X$ may be explicitly added to the set.
Also, it is noted that the modeling of the changes $\Delta {\cal U}^{(n-1)}$ with respect to ${\cal U}^{(n-1)}$ is equivalent to an ML model of the tangent manifold;
this is a direct approach, as opposed to that used in other works \citep{ZLiu-2021,YMototake-2021}.

Data for the variables in $X$ and $Y$ will be used to derive a model of the form $Y = h(X)$, where $h$ generically represents the machine learning model.
The FJet approach does {\em not} restrict the type of ML regression model used; for example,  $h$ may be a neural net (NN), a random forest,
a boosting model, etc.

\subsection*{Methodology}
\label{sec:methodology}

There are several steps to the FJet approach, which are now discussed.

{\bf Step 1}--{\em Create smooth fits to raw data}.
The first step is taking the data provided on a system and putting it into a form usable by the FJet approach.
Note that the data may appear as a time series (i.e., trajectory data) or it may be obtained by random sampling.
Also, in order to obtain derivative information \cite{Knowles-2014,Letellier-2009,Wiener-1950} a local fit to this data is first done, and derivatives are computed with respect to that.
However, in the examples to be shown, the update and derivative information is derived from synthetic data, obviating this step.

{\bf Step 2}--{\em Data for updates}.
The second step is to use the smooth fits created in step \#1 to obtain values for changes in $u$, $\dot{u}$ over a fixed time step $\epsilon$.
In the case of a $2$nd-order ODE, they are simply defined as
\begin{subequations}
\begin{align}
\Delta u & = u(t + \epsilon ) - u(t) \\
\Delta \dot{u} & = \dot{u}(t + \epsilon ) - \dot{u}(t) \, .
\end{align}
\label{eqn:Delta_defn}
\end{subequations}
At this point the data is in the form $(X,Y)$, which for a $2$nd-order autonomous ODE would appear as $(u, \dot{u}, \Delta u, \Delta\dot{u})$ for each time $t$.
Also, note that for an $n$th-order ODE, the set of response variables would include up to $\Delta u^{ (n-1) }$.

{\bf Step 3}--{\em ML model $h$}.
The generic formula for modeling the system as $Y = h(X)$ must be specified for individual elements of $Y$.
For the $2$nd-order example it is written as $h = (h_1,h_2)^T$, which are defined as
\begin{subequations}
\begin{align}
\Delta u & = h_1 (u,\dot{u}) \\
\Delta \dot{u} & = h_2 (u,\dot{u}) \, .
\end{align}
\label{eqn:h_model}
\end{subequations}
As mentioned previously, any suitable ML model may be used for $h_1$ or $h_2$, such as NNs, random forest, boosting methods, etc.
However, a particularly transparent and natural choice is something here named {\em feature regression}.
It is written as
\begin{align}
h = \alpha_1 f_1 + \alpha_2 f_2 + \cdots
\label{eqn:featurereg}
\end{align}
with scalars $\alpha_i$ and features $f_i$ which are functions of the variables of $X$; 
this can be thought of as an extension of polynomial regression.
In the examples, $f_i$ are typically monomials or may involve trigonometric functions.
The advantage this type of model offers is that it simplifies the interpretation of the model in terms of a DE when the time step $\epsilon \rightarrow 0$.
Other authors refer to this as a {\em linear in the parameter model} \citep{Aguirre-review-2009}.

{\bf Step 4}--{\em Extrapolation}.
Having obtained a model for the updates for a given time step $\epsilon$, it is straightforward to use it to
generate a time-sequence of values.  This is done by the {\em generative algorithm} given in Algo.~\ref{algo:generative}.
\begin{algorithm}[H]
\caption{: Generative Algorithm: $2$nd-order}
\begin{algorithmic}
\STATE{ Input: $t_0$, $N$ }
\STATE{ Init: $t = t_0$ }
\STATE{ Init: $u = u(t)$ }
\STATE{ Init: $\dot{u} = \dot{u}(t)$ }
\FOR{ i = $0$ to $N$ }
\STATE{ $\text{tmp\_}h_1  = h_1(u,\dot{u})  $ }
\STATE{ $	\text{tmp\_}h_2 = h_2(u,\dot{u})  $ }
\STATE{ $	u \leftarrow u + \text{tmp\_}h_1 $ }
\STATE{ $	\dot{u} \leftarrow \dot{u} + \text{tmp\_}h_2 $ }
\STATE{ $t \leftarrow t + \epsilon$ }
\STATE{ print($u,t$) }
\ENDFOR
\end{algorithmic}
\label{algo:generative}
\end{algorithm}
Note that as written, this algorithm applies to the $2$nd-order example where $Y = \{ \Delta u, \Delta \dot{u} \}$ and $X = \{ u, \dot{u} \}$.

{\bf Step 5}--{\em Refine model}.
Recall that the determination of the model $h$ in Step \#3 is based on a fit of the response ($Y$) variables as a function of the predictors ($X$).
It is not based on how well the values for $u(t)$ it generates match those of the data.
Thus, this presents an opportunity for a second optimization, in which the parameters obtained for $h$ in Step \#3 can be fine-tuned
to better match the raw data.
This is discussed in greater detail in Appendix~\ref{sec:app-opt}.

An additional step that can be taken to improve the results is to limit the domain of training data
to only encompass the region of phase space where extrapolations are done.
Doing otherwise will bias the model on unexplored regions, reducing the quality of the results in the domain of interest.

{\bf Step 6}--{\em Stability/accuracy}.
In this paper, the accuracy will be in part assessed through a comparison to the best solution available.
In one instance this will be the exact solution; in the others it will be to an extrapolation using RK4.
When available, comparisons will also be made to the energy of the system.

A potential  limiting factor to the stability of the FJet solution is the completeness of the feature set.
It is shown in Appendix~\ref{sec:app-RK} that the feature set is implied by an expansion of the Runge-Kutta numerical integration scheme, to a given order in the time step.
This will be confirmed in the examples, where FJet achieves a level of accuracy of (at least) RK4 in one example,
and a level of accuracy comparable to RK2 in the other two.

Finally, the deviation of $u$, $\dot{u}$ and parameter values measured with respect to their expected values, will be evaluated with the function
\begin{equation}
{\cal E}_\sigma (p,p_{\text{ref}} ) = \ log_{10} | p - p_{\text{ref}} | \, .
\label{eqn:param-error}
\end{equation}
Here $p$ is the value at hand (of $u$, $\dot{u}$ or a parameter), $p_{\text{ref}}$ is the reference value, and $\sigma$ indicates the level of noise.
This measure will be used throughout the paper.




{\bf Step 7}--{\em Underlying DE.}
What has been discussed thus far is how to determine an ML model $h$ for one value of $\epsilon$.
The next step is repeating the entire process for different values of $\epsilon$.
This will allow one to determine a fitted $\epsilon$-dependence of the parameters for the model $h$, and from that 
extrapolate their dependence to $\epsilon=0$.
It is in this $\epsilon = 0$ limit that comparisons can be made to differential quantities, ultmately replacing
$\epsilon$ with $dt$, $\Delta u$ with $du$, and $\Delta \dot{u}$ with $d\dot{u}$.
From this, a derivation of the underlying DE immediately follows.
Note that this is manifestly more robust than just using a single value of $\epsilon$, which has been the standard approach by other authors (e.g., \citep{Crutchfield-1987,Rudy-2017,Aguirre-review-2009}).

{\bf Step 8}--{\em Self-consistency \& uncertainty quantification}.
Beginning from the DE found in the previous step, a numerical integration scheme for it can be given (e.g., RK), 
and that in turn can be Taylor series expanded with respect to its time step ($\epsilon$).
After grouping the terms by $\epsilon$, the collection of functions for a given order in $\epsilon$ are identified as the feature set $X_n$ that can be used to fit the data with an error estimate of ${\cal O}(\epsilon^n)$.

\begin{figure}
\includegraphics[scale=0.32]{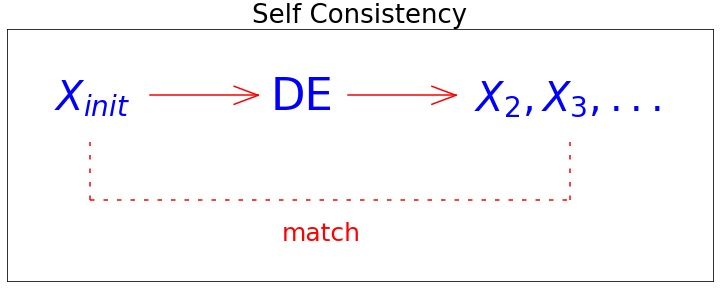}
\caption{
This figure is meant to illustrate how an initial set of features ($X_{init)}$ leads to a DE, which in turn leads to a set of features ($X_n$) which are each
associated with an error estimate of ${\cal O}(\epsilon^n)$.
Having gone through these steps, it is suggested that one may then repeat the process beginning with one of the $X_n$
to see if it leads to the same DE.  If it does, then it may be said to be {\em self-consistent}.
}
\label{fig:fcn_space}
\end{figure}

Defining $X_{init}$ as the initial feature set used in the modeling (cf. Step \#3), there is now an opportunity for a self-consistency check by comparing $X_{init}$ to an $X_n$.
Choosing $n$ such that $X_n$ is similar to $X_{init}$, and then repeating the previous steps with $X_n$, it should then be possible to recover the same DE as already found;
if so, it can be said to be self-consistent.
Fig.~\ref{fig:fcn_space} is given to illustrate this matching of $X_{init}$ with one of the $X_n$.
Further discussion of $X_n$ can be found in Sec.~\ref{sec:fcnspace}.

Finally, in an actual application, it's important to know the uncertainty quantification;
otherwise the method is really just a proof of concept.
The reader should note that the standard way of doing this type of modeling (e.g., \citep{Crutchfield-1987,Rudy-2017,Aguirre-review-2009}) has only involved an ad hoc selection of features $X$, and included no uncertainty quantification.

\subsubsection*{Training Data}
\label{sec:trainingdata}

The training data is synthetic and will be generated by evolving the known, underlying DE forward in time by an amount $\epsilon$;
this will be done for randomly chosen values of $u$ and $\dot{u}$.  The set of $\epsilon$ values to be used below will range from $0.001$ to $0.1$.
To mitigate unnecessary discretization errors, the DE will be generated by a number of iterations of a time step $\epsilon_{base} = 0.001$.
For example, to obtain data for a single instance of $(\Delta u, \Delta \dot{u})$ at $\epsilon = 0.1$, the RK4 scheme will be iterated 100 times using $\epsilon_{base}$.

In addition, the bulk of the examples used training data that was made more realistic by including an additive {\em measurement error} of $\sigma \epsilon \nu$ 
to each entry of $u$ and $\dot{u}$
\begin{align*}
u & \rightarrow u + \sigma \epsilon \nu \\
\dot{u} & \rightarrow \dot{u} + \sigma \epsilon \nu'  
\end{align*}
where $\nu$ and $\nu'$ represent independent draws from a zero-mean, unit-variance normal distribution.
Note that this also impacts functions of $u$ and $\dot{u}$, as well as $\Delta u$ and $\Delta \dot{u}$.
The time step $\epsilon$ was included as a factor in the strength of $\nu$ to reflect that these measurements were made with respect to a smoothed fit of $u$,
as discussed in Sec.~\ref{sec:methodology}; a noise coefficient with constant strength would overwhelm $\Delta u$ and $\Delta \dot{u}$ for small $\epsilon$.
Systematic deviations for $u$, $\dot{u}$ that do not vanish as $\epsilon \rightarrow 0$ will not be considered.
In this paper, the data will typically be modeled with $\sigma = 0$, $0.1$ and $0.2$.

\section{Example: Harmonic Oscillator}
\label{sec:HO}

The equation for a harmonic oscillator is relevant for a myriad of physical systems,
such as the small angle limit of a pendulum.
Here, $u$ represents the deviation from the minimum, $\omega_0$ its natural frequency, and $\gamma$ its damping coefficient
\begin{equation}
\ddot{u} + 2\gamma \dot{u} + \omega_0^2 u = 0 \, .
\label{eqn:HO}
\end{equation}
This section will only consider the underdamped case, unless stated otherwise.

\subsection*{Training Data}

Training data was generated from 2000 random samples,
using $\omega_0=1$, $\gamma = 0.1$, $T=2\pi/\omega_0$, $t \in (0,2T)$,  $u \in (-2,2)$, and $\dot{u} \in (-2,2)$.
In order to generate the $\Delta u$ and $\Delta \dot{u}$, the data points were created with a time step $\epsilon=0.1$ (with iterated updates of RK4) and $\sigma = 0.2$,
as described near the end of Sec.~\ref{sec:trainingdata}.

\begin{figure}
\includegraphics[scale=0.18]{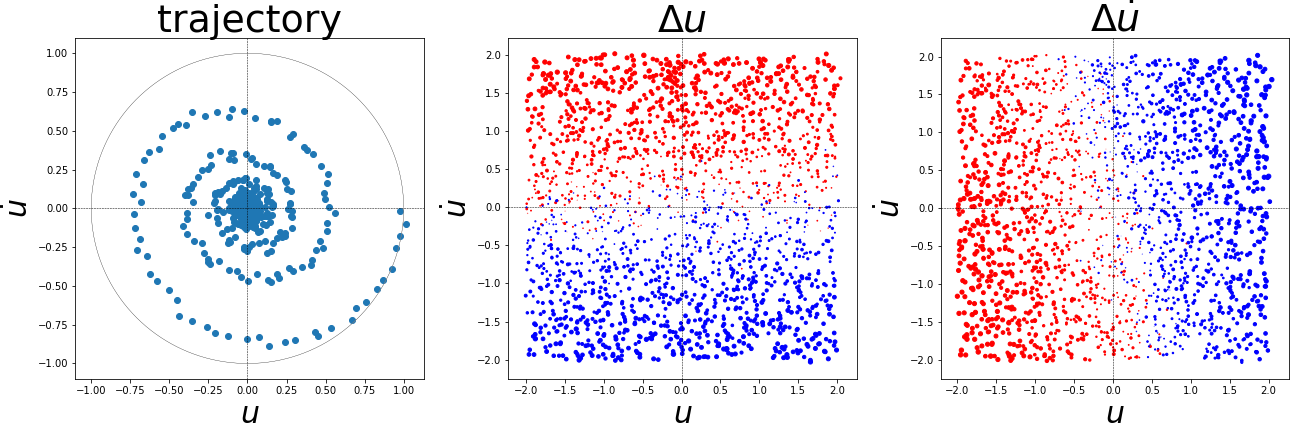}
\caption{
Three plots for the harmonic oscillator, using a noise level of $\sigma=0.2$.  
The left plot corresponds to the (clockwise) path taken by updates of the system, starting from an initial condition of $(u,\dot{u})=(1,0)$.  
The light gray circle is added as a visual reference, and corresponds to the path that would have been taken in the undamped, noiseless case.  
The center and right plots correspond to the changes in $u$ and $\dot{u}$, as defined in Eq.~\ref{eqn:Delta_defn}.
}
\label{fig:HO_Deltas}
\end{figure}

In the left plot of Fig.~\ref{fig:HO_Deltas} a typical trajectory is shown using $\sigma=0.2$, to show the impact of the noise.
In both the center and right plots, the differences ($\Delta u$, $\Delta \dot{u}$) (computed between times $t$ and $t+\epsilon$) are plotted with respect to the values at time $t$.
In the plots, the color of the points is determined by the sign of $\Delta u$ or $\Delta \dot{u}$: it is red (blue) if the sign is positive (negative).
Also, the size of the dots is proportional to the magnitude of $\Delta u$ (or $\Delta \dot{u}$); the same scaling was used for both plots in the figure.
These figures provide useful information as to how to model $\Delta u$ and $\Delta \dot{u}$ over this space.
They demonstrate how to capture the dynamic information in static plots.

\subsection*{Model}

Given these observations, it appears that it would suffice from a modeling perspective to simply have $h_1$ and $h_2$ be 
linear combinations of $u$ and $\dot{u}$.
However, following the guidance from Eq.~\ref{eqn:rubric}, a larger class of candidate features 
might instead be considered
\begin{align*}
\{ u, \dot{u}, t, t^2, u^2, u\dot{u}, \dots \} \, .
\end{align*}
It will be understood from Appendix A that several of these terms need not be considered (such as $t$, $t^2$) for dynamics with no explicit time dependence.
(No constant term is expected, since it appears the dot size is zero near the origin, in both plots in Fig.~\ref{fig:HO_Deltas}.
Also, note that if $\ddot{u}$ were included in the predictors, a collinearity would manifest itself, a signal to the modeler that it should be removed.)

As already discussed, while many ML models can be used (such as a NN), it is convenient here to use {\em feature regression}, since it will later facilitate determining the underlying DE.
From this set of candidate features, the usual techniques of regression \citep{Hastie_ESL2} can be used to remove irrelevant predictors.
The resulting response ($Y$) and predictor ($X$) variables are then chosen to be
\begin{align}
Y & = ( \Delta u, \Delta \dot{u} )^T \\
X & = ( u, \dot{u} )^T  
\end{align}
and the feature regression model $Y = h(X)$ appears as a matrix-vector product
\begin{align}
Y
& =
\epsilon
\begin{bmatrix}
a_1 & a_2  \\
b_1 & b_2 
\end{bmatrix} 
X \, .
\label{eqn:HO_mapping}
\end{align}
In this approach, one can now determine the coefficients $a_i$ and $b_i$ ($i=1,2$) by fitting it to the training data using $\epsilon=0.1$.
Note that an explicit regularization was not used, in contrast to approaches by other authors \citep[e.g.,][]{Crutchfield-1987,Rudy-2017}.
With regularization, one might have used, for example, elastic net regularization \citep{Hastie_ESL2} to drive small, less important coefficients down to zero.
Instead, it will be shown next that creating a feature regression model for {\em several} values of $\epsilon$, 
with no explicit regularization, can lead to a robust, controlled derivation of the underlying DE.


\subsection*{Underlying DE}

\begin{figure}
\includegraphics[scale=0.21]{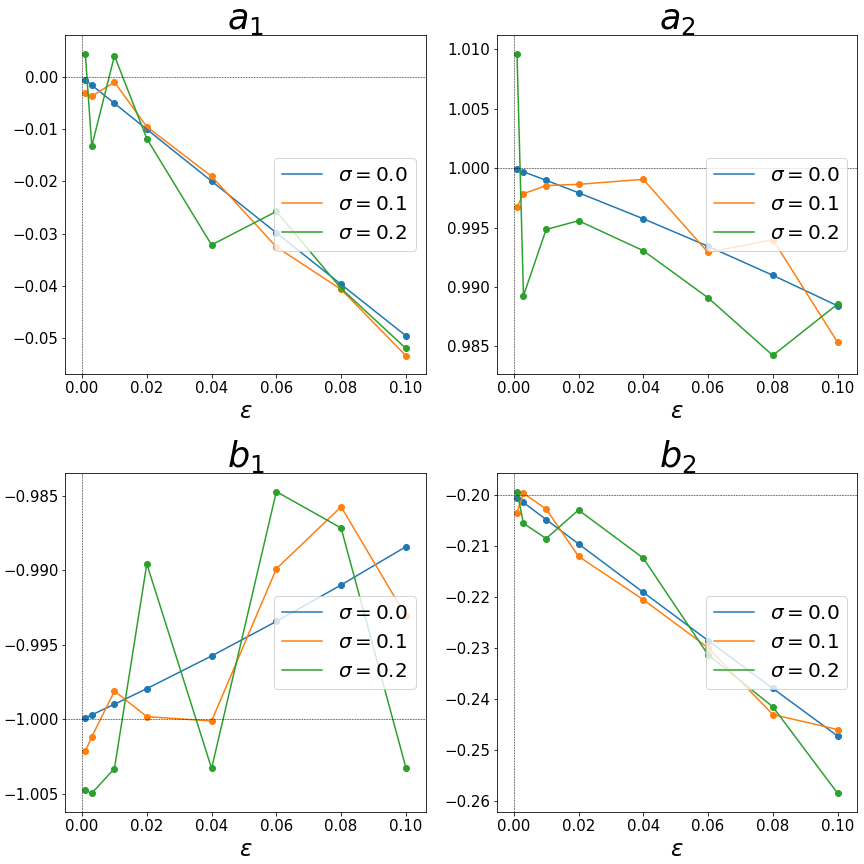}
\caption{Plots of the $\epsilon$-dependence of the coefficients in the model $h(X)$
(cf. Eq.~\ref{eqn:HO_mapping}).  
The intersection of the thin vertical and horizontal gray lines indicate the correct value at $\epsilon=0$.  Note the different scales for the error for each plot.}
\label{fig:HO_DE_params}
\end{figure}

The process of fitting $Y$ to $X$ for a single $\epsilon$ is now repeated for the set of $\epsilon$-values  $\{0.001, ...0.1\}$,
leading to the plots in  Fig.~\ref{fig:HO_DE_params}; these plots include data for the cases $\sigma=0, 0.1, \text{and } 0.2$.
Best line fits were used to model the $\epsilon$-dependence, with the results collected in Table~\ref{tab:fits} in Appendix~\ref{sec:app-params}.
These fits were extrapolated to $\epsilon =0$ to determine parameter values for the underlying DE.
Equation~\ref{eqn:HO_mapping} may now be rewritten in this limiting case, and the variables will be written using the standard differential notation 
(i.e., $\epsilon \rightarrow dt$, $\Delta u \rightarrow du$, $\Delta \dot{u} \rightarrow d\dot{u}$ ), producing
\begin{subequations}
\begin{align}
du & = dt \, [ \dot{u} ] \label{eqn:HO_first} \\
d\dot{u} & = dt \, [  -0.2 \dot{u} -u ]  \, .  \label{eqn:HO_second}
\end{align}
\end{subequations}
(Of course, this is a slight abuse of notation, since derivatives are normally defined in a limiting procedure.)
The first of these equations (Eq.~\ref{eqn:HO_first}) is just a contact condition, and reflects the fact that in the space $(u,\dot{u})$, not all three
quantities $du$, $dt$, $\dot{u}$ are independent (i.e., $\dot{u} = du/dt$).
Finally, when Eq.~\ref{eqn:HO_first} and Eq.~\ref{eqn:HO_second} are combined, they are just equivalent to the original equation of motion for the damped harmonic oscillator (Eq.~\ref{eqn:HO}),
with the chosen parameter values ($\omega_0=1$ and $\gamma = 0.1$).

\subsection{Parameter Dependence}

What was obtained in the previous section is the equation of motion based on the choice $\omega_0=1$, $\gamma=0.1$.
In this section, the actual functional dependence on $\omega_0$ and $\gamma$ within the equation of motion is deduced.

\begin{figure}
\includegraphics[scale=0.22]{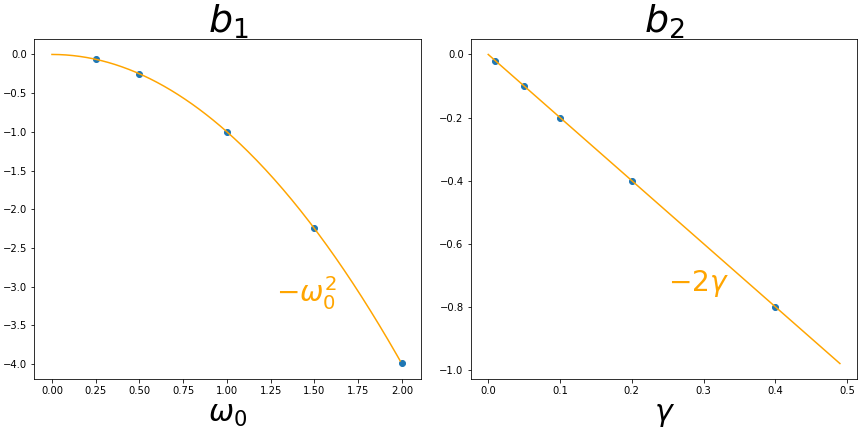}
\caption{The dependence of $b_1$ and $b_2$ as a function of $\omega_0$ and $\gamma$ is plotted using blue dots.  As a visual guide, the exact curves for $-\omega_0^2$ and $-2\gamma$ are added as orange curves to the left and right plots, respectively.}
\label{fig:HO_param_combo}
\end{figure}

The idea here is to consider a given value of $\epsilon$, and several values of $\omega_0$.
This produces a plot versus $\omega_0$, which is then fit with a low-order polynomial in $\omega_0$.
Thus, the fitting parameters are now determined for that one, given value of $\epsilon$.
Next, this procedure is repeated for several other small values of $\epsilon$, and the result
is extrapolated to $\epsilon=0$.  
In this section the fitting was done for the zero noise case (i.e., $\sigma = 0$), and led to the results:
$b_1 = - \omega_0^2 + \dots$, $b_2 = - 2 \gamma + \dots$.
Shown in Fig.~\ref{fig:HO_param_combo} is the $\omega_0$ and $\gamma$ dependence for these parameters in the specific case of $\epsilon=0.001$.
From this, Eq.~\ref{eqn:HO_first} and Eq.~\ref{eqn:HO_second} can be replaced with
\begin{align}
du & = dt \, [ \dot{u} ] \\
d\dot{u} & = dt \, [ - 2\gamma \dot{u} - \omega_0^2 u ]
\end{align}
from which the original equation of motion (Eq.~\ref{eqn:HO}) immediately follows.
In summary, it has been shown how to obtain the complete equation of motion (along with parameter dependencies), from data.

\subsection*{Extrapolation \& Stability}

There are three approaches for this example in which the stability and the extrapolation accuracy of the FJet models can be evaluated.

\begin{figure}
\includegraphics[scale=0.35]{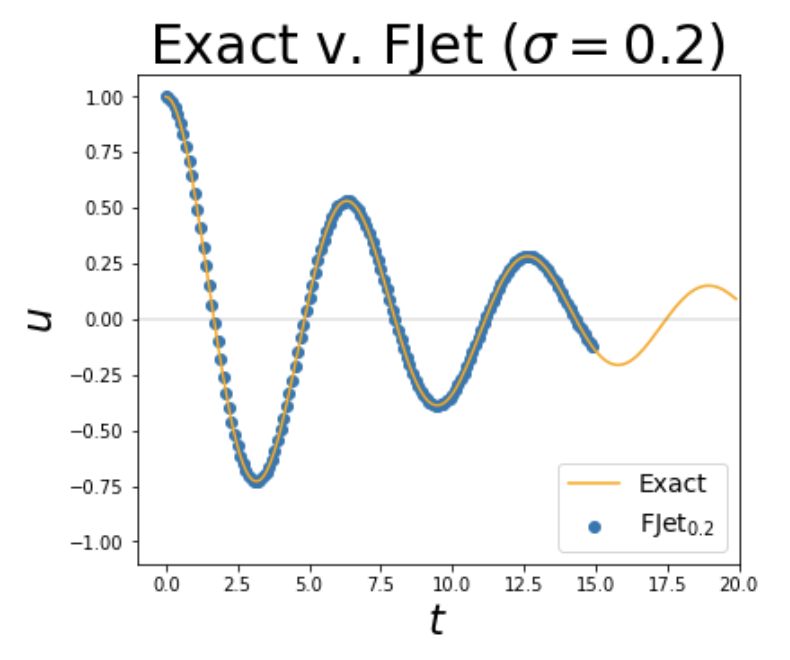}
\caption{A comparison of the extrapolation of the exact solution to that generated by FJet with $\sigma = 0.2$.
}
\label{fig:HO-phase-space}
\end{figure}

{\em First Approach:} One may do an extrapolation in time using the model parameter values for FJet with $\sigma = 0.2$,
and do a visual comparison.
Upon using the generative algorithm with $\epsilon=0.1$, Fig.~\ref{fig:HO-phase-space} results.
Thus for these shorter times, close agreement is visually confirmed.
However, for longer times such a graph is less informative, and so the next approach is considered.

\begin{figure}
\includegraphics[scale=0.25]{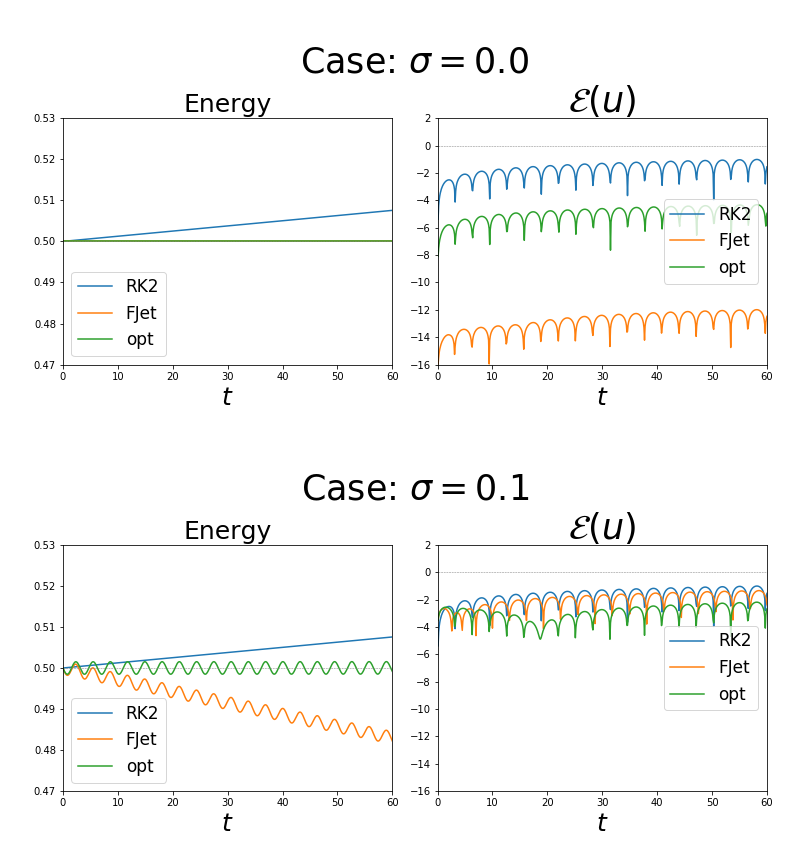}
\caption{Shown are plots of the errors for the damped harmonic oscillator.
For $\sigma=0$, the energy values for {\em opt} overlaps that of FJet, and neither has any discernable deviation from the exact value of 0.5.  Also, for this case note the much lower error values for FJet on this log scale; the optimization process (Appendix~\ref{sec:app-opt}) reduces the accuracy from this nearly ideal case.  For the case $\sigma = 0.1$, the {\em opt} solution fares better than FJet or RK2.}
\label{fig:HO-extrap-errors-00}
\end{figure}

{\em Second Approach:} A way to understand the extrapolation accuracy is to compare it to the exact solution mentioned earlier.
The error measure defined in Eq.~\ref{eqn:param-error} is used in Fig.~\ref{fig:HO-extrap-errors-00} to compare the magnitudes of the errors for the solutions found using RK and FJet; included in the plot are the results from an optimized solution (using Appendix~\ref{sec:app-opt}).
What is most noticeable in the $\sigma=0$ plot is the small error of FJet$_0$ compared to RK2 and the optimized solution; note the log scale.

{\em Third Approach:} A way to compare the accuracy of extrapolations is to compute the energy in the undamped case: 
$E = \frac{1}{2} ( u^2 + \dot{u}^2 )$, where now $\omega_0=1$ and $\gamma=0$.
Following \cite{Young_notes}, the energy at the $n$th iteration ($E_n$) is parametrized as exponentially changing 
\begin{align}
E_n / E_0 & = \exp( \lambda n)  \, .
\end{align}
Thus, a $\lambda$ closer to $0$ means it is more stable.
Iterating for $10^4$ iterations, with the initial condition of $(u,\dot{u})=(1,0)$, led to the values in Table \ref{tab:stability-logrsq}.
\begin{table}[b]
\caption{ \label{tab:stability-logrsq}
In the case where $\gamma = 0$ (undamped), and $\sigma=0$ (zero noise), FJet (using feature regression) remains stable about $10^9$ times longer than RK4.
}
\begin{ruledtabular}
\begin{tabular}{lc}
  \textrm{scheme}&
  $\lambda$ \\
\colrule
Euler &    $9.94 \times 10^{-2}$ \\
RK2 &    $2.49 \times 10^{-5}$ \\
RK4 &    $-1.38 \times 10^{-8}$ \\
FJet$_0$ &    $-1.63 \times 10^{-17}$ \\
\end{tabular}
\end{ruledtabular}
\end{table}
The most obvious aspect to this table is the much smaller value of $\lambda$ for FJet$_0$ (with feature regression);
it means that FJet$_0$ remains stable for about $10^9$ times longer compared to RK4.
The parameter values used for FJet$_0$ were computed using $\epsilon = 0.1$ and appear in Table.~\ref{tab:a1a2expt}, along with comparison values
from Euler/RK2/RK4.  
\begin{table}[b]
\caption{Parameter values for the {\em undamped} harmonic oscillator with $\gamma=0$ and $\epsilon=0.1$.
The third column ``FJet$_0$" refers to the zero-noise case, and was found using feature regression.}
\begin{ruledtabular}
\begin{tabular}{lcccc}
\textbf{}  &   Euler  &   RK2  &  RK4  &  FJet$_0$  \\
 $a_1$  &  $0$     &     $-0.005$  &  $-0.00499583\bar{3}$  &  $-0.004995834721974124$ \\
 $a_2$  &  $0.1$  &      $0.1$      &   $0.09983\bar{3}$      &   $0.09983341664682731$ 
\end{tabular}
\end{ruledtabular}
\label{tab:a1a2expt}
\end{table}


\section{Example: Pendulum}
\label{sec:Pendulum}

The equation of motion for a damped pendulum is
\begin{align}
\ddot{u} + 2\gamma\dot{u} + \omega_0^2 \sin u =  0 
\label{eqn:pendulum} 
\end{align}
where $\omega_0$ is the natural frequency in the small-$u$ limit and $\gamma$ is the damping coefficient.

\subsection*{Training Data}

\begin{figure}
\includegraphics[scale=0.18]{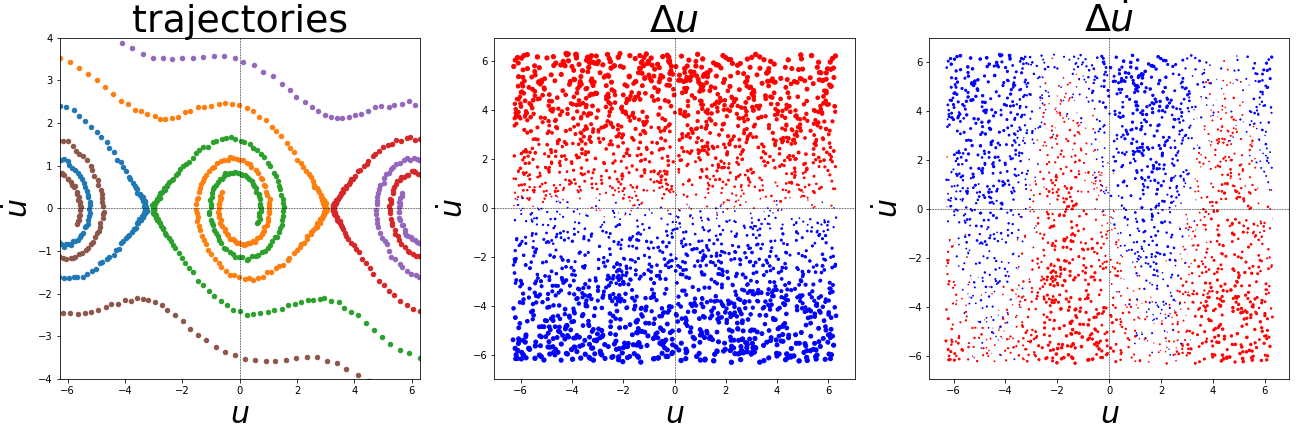}
\caption{
Three plots for the pendulum, using a noise level of $\sigma=0.2$.  
The left plot displays multiple trajectories, starting from a number of different initial conditions.
Note the $2\pi$ periodicity along $u$.
The center and right plots correspond to the changes in $u$ and $\dot{u}$, as defined in Eq.~\ref{eqn:Delta_defn}.
}
\label{fig:pendulum_Deltas}
\end{figure}

The parameter values $\omega_0=1.0$ and $\gamma=0.1$ are used in Fig.~\ref{fig:pendulum_Deltas},
which includes multi-start trajectories (left plot), and the update maps $\Delta u$ and $\Delta \dot{u}$ (center and right plots).
Note that in the left plot, the rotation is clockwise; in the upper (lower) half of the plane, the motion is to the right (left).
The $\Delta$-plots were created by randomly selecting 2000 $(u,\dot{u})$ values and then doing a single time step update with RK4 (with iterated updates), 
using $\epsilon=0.1$.
The three plots in this figure have $u$ and $\dot{u}$ each range over $(-2\pi,2\pi)$ to display the periodic structure,
but in the actual training data only the range of $(-\pi,\pi)$ was used.
This created a data pair, and allowed a creation of data for $\Delta u$ and $\Delta \dot{u}$.
It is these plots which will be fitted by an ML procedure.
As before, the dot sizes are proportional to the magnitude of what is being plotted; they are colored red (blue) for positive (negative) values.

\subsection*{Model}

As a starting assumption, based on experience with the harmonic oscillator, one might begin by considering the feature set
$\{ u, \dot{u} \}$.  However, the domain knowledge in this case is that $u$ must be periodic, so instead the starting feature set
under consideration is $\{ \dot{u}, \cos u, \sin u \}$.
As discussed in Appendix~\ref{sec:app-RK}, 
numerical integration techniques suggest one should use a larger feature set, such as
\begin{align*}
\{ \dot{u}, \cos u, \sin u, \dot{u} \cos u, \dot{u}\sin u, \dots \} .
\end{align*}
The elements in this new set may now be taken as input features into an appropriate ML model.
As already discussed, while many ML models can be used (such as NNs), it is convenient here to use {\em feature regression}, since it will facilitate determining the underlying DE.
After following the usual steps in regression \cite{Hastie_ESL2},
the resulting response ($Y$) and predictor ($X$) variables are chosen to be
\begin{subequations}
\begin{align}
Y & = ( \Delta u, \Delta \dot{u} )^T  \label{eqn:Pendulum_Y} \\
X & = ( \dot{u}, \sin u, \dot{u} \cos u )^T  \label{eqn:Pendulum_X} 
\end{align}
\end{subequations}
and the feature regression model $h(X)$ appears as a matrix-vector product,
\begin{align}
Y
& =
\epsilon
\begin{bmatrix}
a_1 & a_2 & a_3 \\
b_1 & b_2 & b_3 \\
\end{bmatrix} 
X \, .
\label{eqn:Pendulum_mapping}
\end{align}
In this approach, one can now determine the coefficients $a_i$ and $b_i$ ($i=1,2,3$) by fitting it to the training data using $\epsilon=0.1$.
(Note that validation and test data could be used for a refined approach.)
In the feature regression modeling, regularization was not used, in contrast to approaches by other authors \citep[e.g.,][]{Crutchfield-1987,Rudy-2017}.

\subsection*{Underlying DE}

\begin{figure}
\includegraphics[scale=0.18]{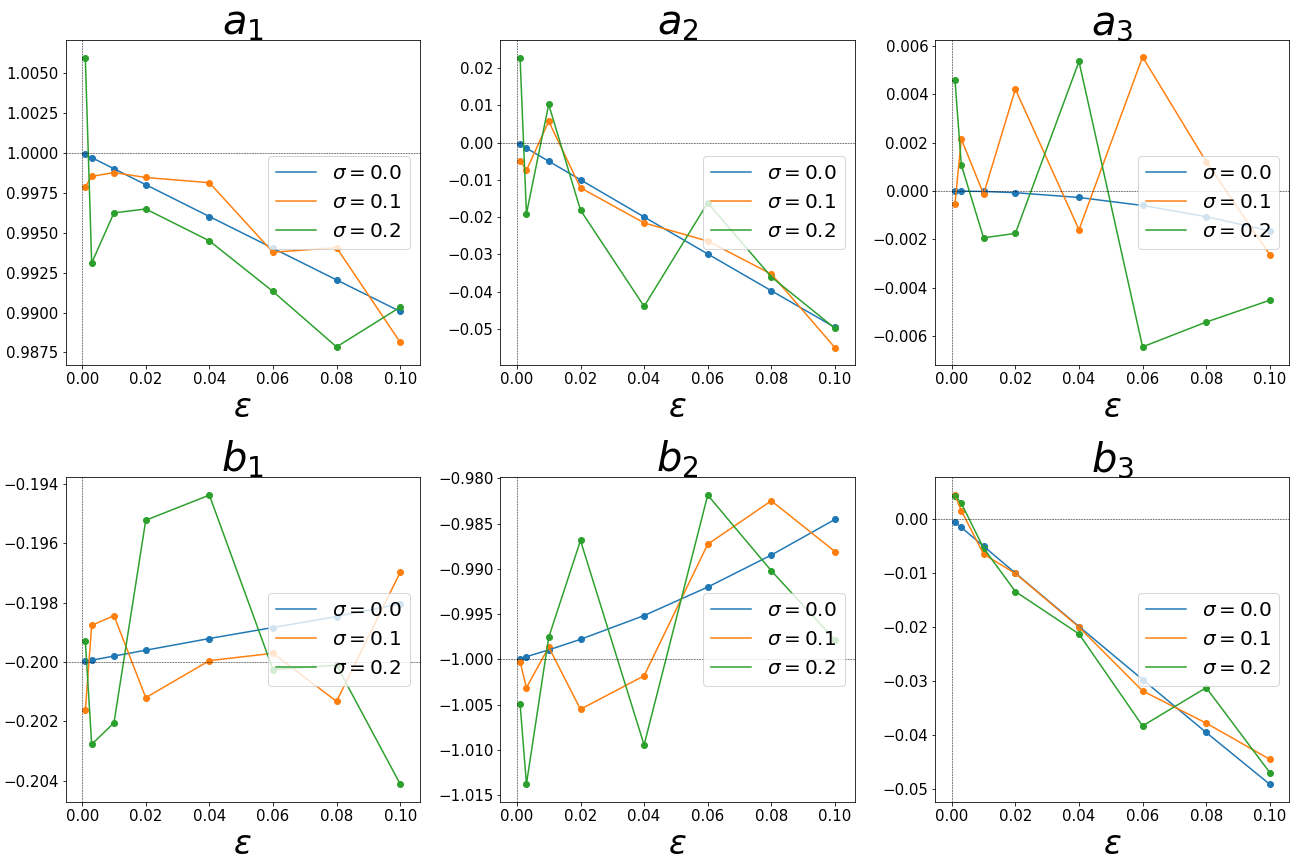}
\caption{Plots of the $\epsilon$-dependence of the coefficients in the model $h(X)$ (cf. Eq.~\ref{eqn:Pendulum_mapping}).
The intersection of the thin vertical and horizontal gray lines indicate the correct value at $\epsilon = 0$.}
\label{fig:Pendulum_DE_params}
\end{figure}

The process of fitting $Y$ to $X$ for a single $\epsilon$ is now repeated for the set of $\epsilon$-values  $\{0.001, ...0.1\}$,
leading to the plots in Fig.~\ref{fig:Pendulum_DE_params}; these plots include data for the cases $\sigma=0$, $\sigma = 0.1$ and $\sigma = 0.2$.
Best line fits were used to model the $\epsilon$-dependence, with the results collected in Table~\ref{tab:fits} in Appendix~\ref{sec:app-params}.
These fits were extrapolated to $\epsilon = 0$ to determine parameter values for the underlying DE.
Equation~\ref{eqn:Pendulum_mapping} may now be rewritten in this limiting case, and the variables will be written using the standard differential notation 
(i.e., $\epsilon \rightarrow dt$, $\Delta u \rightarrow du$, $\Delta \dot{u} \rightarrow d\dot{u}$ ), producing
\begin{subequations}
\begin{align}
du & = dt [ \dot{u} ] \label{eqn:Pendulum_first} \\
d\dot{u} & = dt [  -0.2 \dot{u} -\sin u ] \, . \label{eqn:Pendulum_second}
\end{align}
\end{subequations}
The first of these equations (Eq.~\ref{eqn:Pendulum_first}) is just a contact condition, and reflects the fact that in the space $(u,\dot{u})$, not all three
quantities $du$, $dt$, $\dot{u}$ are independent (i.e., $\dot{u} = du/dt$).
The combination of these two equations is just equivalent to the original equation of motion for the pendulum (Eq.~\ref{eqn:pendulum}),
with the chosen parameter values.

\subsection*{Extrapolation \& Stability}

\begin{figure}
\includegraphics[scale=0.27]{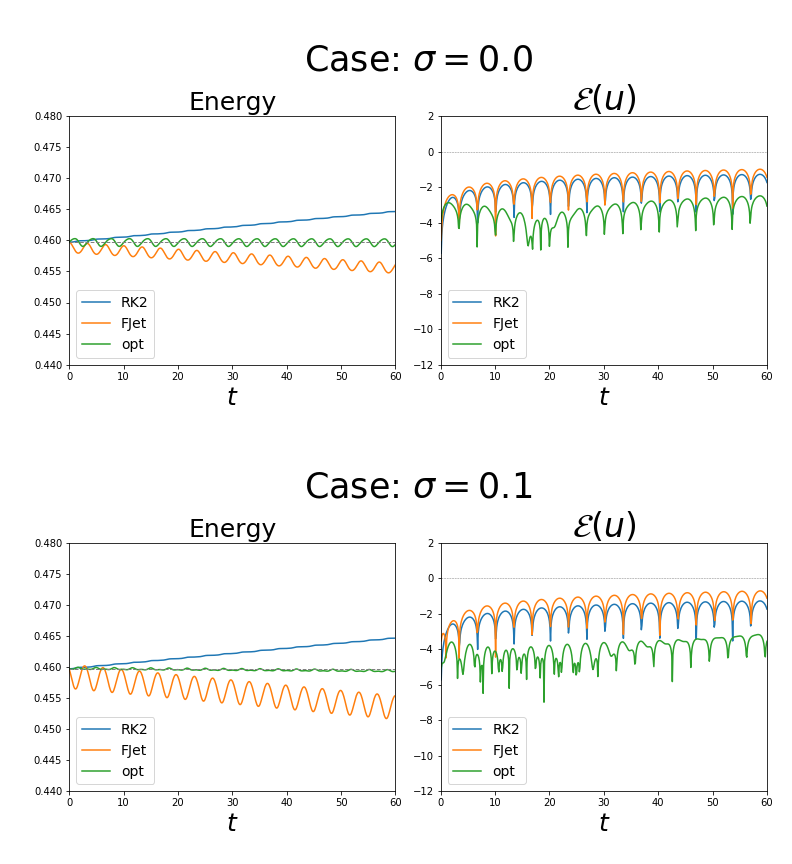}
\caption{
Shown are plots of the errors for the undamped pendulum.
The errors in energy for RK2 and FJet are comparable. 
The {\em opt} solution fares better than FJet or RK2, for either noise level.
Since the initial condition is $(u,\dot{u}) = (1,0)$, the energy is $1-\cos(1) \approx 0.46$.
}
\label{fig:Pendulum_stability_all}
\end{figure}

Using $\epsilon = 0.1$, $\gamma=0$, and the initial conditions $(u,\dot{u})=(1,0)$, values for $a_i$ and $b_i$ were computed similarly as before,
and an extrapolation of $u(t)$ was created using Algo.~\ref{algo:generative}.
This undamped case was used so that the energy ($E$), as defined by
\begin{equation}
E = \frac{1}{2} \dot{u}^2 + \omega_0^2 (1 - \cos u) 
\label{eqn:Pendulum_energy}
\end{equation}
could also be examined to see if it remained conserved.
Shown in Fig.~\ref{fig:Pendulum_stability_all} are plots of the energy and the error (as defined in Eq.~\ref{eqn:param-error}).
The data are from RK2, FJet, and the optimized FJet; this is for the cases of $\sigma=0$ and $0.1$.
As can be seen, the plots reveal that FJet and RK2 perform comparably, while the optimized version of FJet performs much better.

\section{Example: Duffing Oscillator}
\label{sec:Duffing}

The Duffing oscillator \citep{Duffing-1918,Holmes-1976,MoonHolmes-1979,Atlee-book-v1,Jordan-book} is similar to the damped oscillator seen in Sec.~\ref{sec:HO}, 
except it allows for a nonlinear restoring force and an external force $p(t)$.
It appears here as
\begin{subequations}
\begin{align}
& \ddot{u} + 2\gamma\dot{u} + \alpha u + \beta u^3 = p(t)  \label{eqn:duffing} \\
& p(t) = A \cos \Omega t  \, . \label{eqn:duffing_force}
\end{align}
\label{eqn:duffing_both}
\end{subequations}

\subsection*{Training Data}

\begin{figure}
\begin{center}
\includegraphics[scale=0.18]{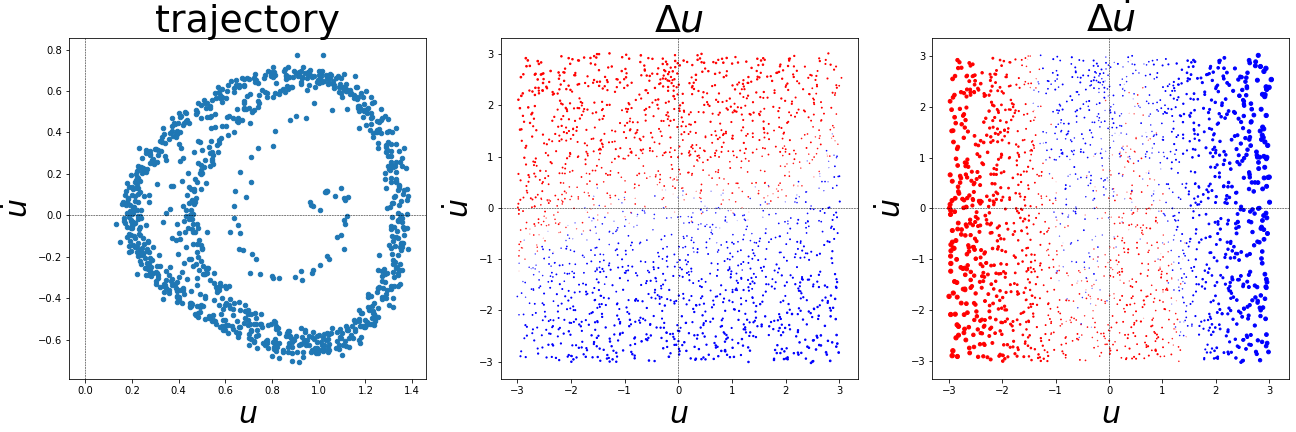}
\caption{
Three plots for the Duffing oscillator, using a noise level of $\sigma=0.2$.  
The left plot corresponds to the (clockwise) path taken by updates of the system, starting from an initial condition of $(u,\dot{u})=(1,0)$.  
The center and right plots correspond to the changes in $u$ and $\dot{u}$, as defined in Eq.~\ref{eqn:Delta_defn}.
Also, note the different scale of the left plot compared to the others.}
\label{fig:Duffing_Deltas}
\end{center}
\end{figure}

This example used the parameter values $\gamma=0.15$, $\alpha= -1$, $\beta = 1$, $A=0.28$, and $\Omega=1.2$,
resulting in the plots of Fig.~\ref{fig:Duffing_Deltas}.
The left plot was created from a single run with initial condition $(u,\dot{u})=(1,0)$; the updates lead to a clockwise rotation.
Note that the orbit displays a period-2 oscillation (cf. pp453-462 in \cite{Jordan-book}, and \cite{Holmes-1976}), 
and that it has only positive $u$-values (since the $\alpha$ and $\beta$ values create a double-well potential).
The center and right plots are the update maps $\Delta u$ and $\Delta \dot{u}$.
As before, the dot sizes are proportional to the magnitude of what is being plotted, and are red (blue) for positive (negative) values.
The training data was obtained by randomly sampling 2000 points; 
the sampling domain for $u$, $\dot{u}$ and $t$ were $(-3,3)$, $(-3,3)$, and $(0,4\pi/\Omega)$, respectively.
As before, the second point was created by doing a single time-step update (with $\epsilon=0.1$) using RK4 with iterated updates.
The random time was inserted into Eq.~\ref{eqn:duffing_force} to create sampling for $p$ and $\dot{p}$.  
This created a data pair, and allowed a creation of $\Delta u$ and $\Delta \dot{u}$.
Note that the plots for $\Delta u$ and $\Delta \dot{u}$ do not include the independent variables $p$ and $\dot{p}$ (or time).
Because of this, Fig.~\ref{fig:Duffing_Deltas} is a plot that has a collapsed domain from $(u,\dot{u},p,\dot{p})$ to $(u,\dot{u})$, and thus there is a slight overlap
of red and blue dots, even for $\sigma=0$.  (Of course, the presence of noise (i.e., $\sigma \neq 0$) also leads to a mixing of the red and blue dots.)
Finally, it is noted that to create data that will lead to models that better support extrapolations with different $p(t)$ (e.g., different values for $A$ or $\Omega$),
it could be advantageous to also sample over a range of $A$ and $\Omega$.

\subsection*{Model}

While a starting assumption for the feature set might be $\{ u, \dot{u}, p \}$,
the discussion of Runge-Kutta in Appendix~\ref{sec:app-RK} suggests it should be enlarged, such as
\begin{eqnarray}
\{ && u, \dot{u}, u^2, u\dot{u}, \dot{u}^2, u^3, u^2 \dot{u}, u \dot{u}^2, \nonumber\\
&& \dot{u}^3, p, \dot{p}, pu, p\dot{u}, \dot{p} u, \dot{p}\dot{u}, \dots \} .
\end{eqnarray}
The elements in this new set may now be taken as input features into an appropriate ML model.
As already discussed, while many ML models can be used (such as a NN), it is convenient here to use {\em feature regression}, since it will facilitate determining the underlying DE.
After following the usual steps in regression \cite{Hastie_ESL2}
the resulting response ($Y$) and predictor ($X$) variables are chosen to be
\begin{align}
Y & = ( \Delta u, \Delta \dot{u} )^T  \label{eqn:Duffing_Y} \\
X & = ( u, \dot{u}, u^3, u^2\dot{u}, u\dot{u}^2, \dot{u}^3, p, \dot{p} )^T  \, . \label{eqn:Duffing_X}
\end{align}
In Appendix~\ref{sec:app-RK}, it is shown how an RK expansion motivates which terms should be included, 
and the terms $u\dot{u}^2$ and $\dot{u}^3$ were not among them.
However, here they were included to demonstrate that the technique is resilient against the addition of unnecessary terms.
The feature regression model $Y = h(X)$ appears as a matrix-vector product,
\begin{align}
Y
& =
\epsilon
\begin{bmatrix}
a_1 & a_2 & a_3 & 0 & 0 & 0 & a_7 & 0 \\
b_1 & b_2 & b_3 & b_4 & b_5 & b_6 & b_7 & b_8 \\
\end{bmatrix} 
X \, .
\label{eqn:Duffing_mapping}
\end{align}
In this approach, one can now determine the coefficients $a_i$ and $b_i$ ($i=1,...,8$) by fitting it to the training data;
$a_4$, $a_5$, $a_6$, $a_8$ were already identified as being essentially 0.
Note that the resulting FJet mapping can then make use of any forcing term in an extrapolation algorithm;
all that is needed is knowledge of $p$ and $\dot{p}$ at a series of times.
Because of its reliance on a general $p$ (and $\dot{p}$), it resembles a Green's function approach.

In the feature regression modeling, regularization was not used, in contrast to approaches by other authors \citep[e.g.,][]{Crutchfield-1987,Rudy-2017}.
For example, one might have used regularization to drive small, less important coefficients down to zero,
enabling an easier identification of a ``minimal model".
Instead, the $\epsilon$-dependence of these parameters will be studied in the next sub-section, to determine their behavior as $\epsilon \rightarrow 0$ and ascertain the underlying DE.

A model for the Duffing equation was also reconstructed from phase space data by \citep{Menard-2000}, where they included two first-order equations for $p(t)$ to model its oscillatory dynamics.  In doing so, they were able to study the Duffing oscillator as an autonomous system with four first-order ODEs.

\subsection*{Underlying DE}

\begin{figure*}
\includegraphics[scale=0.2]{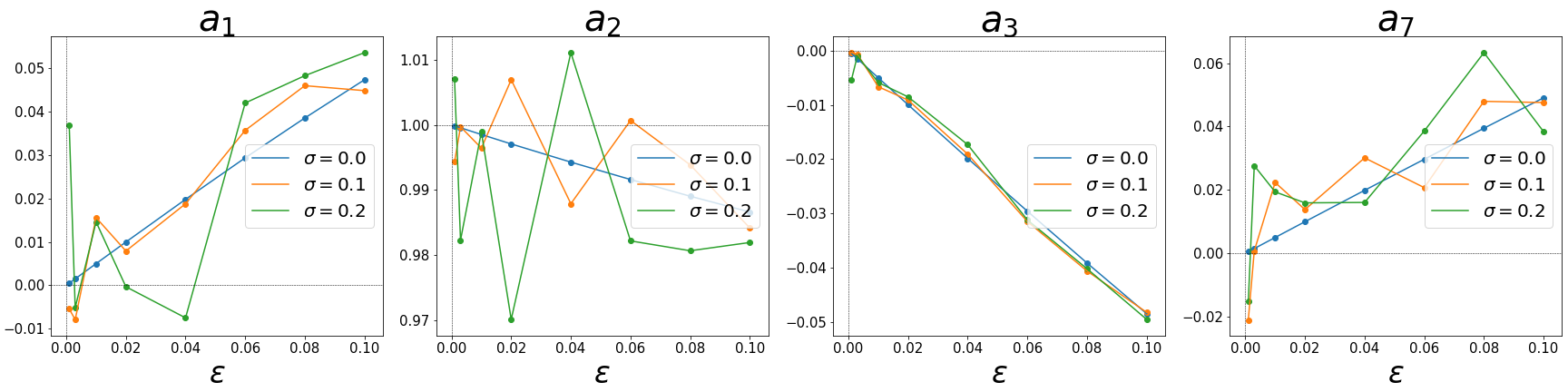}
\caption{Plots of the $\epsilon$-dependence of the $a_i$ coefficients ($i=1,2,3,7$) in the model $h(X)$ (cf. Eq.~\ref{eqn:Duffing_mapping}).
The intersection of the thin vertical and horizontal gray lines indicate the correct value at $\epsilon = 0$. }
\label{fig:Duffing_DE_params_a}
\end{figure*}

\begin{figure*}
\includegraphics[scale=0.2]{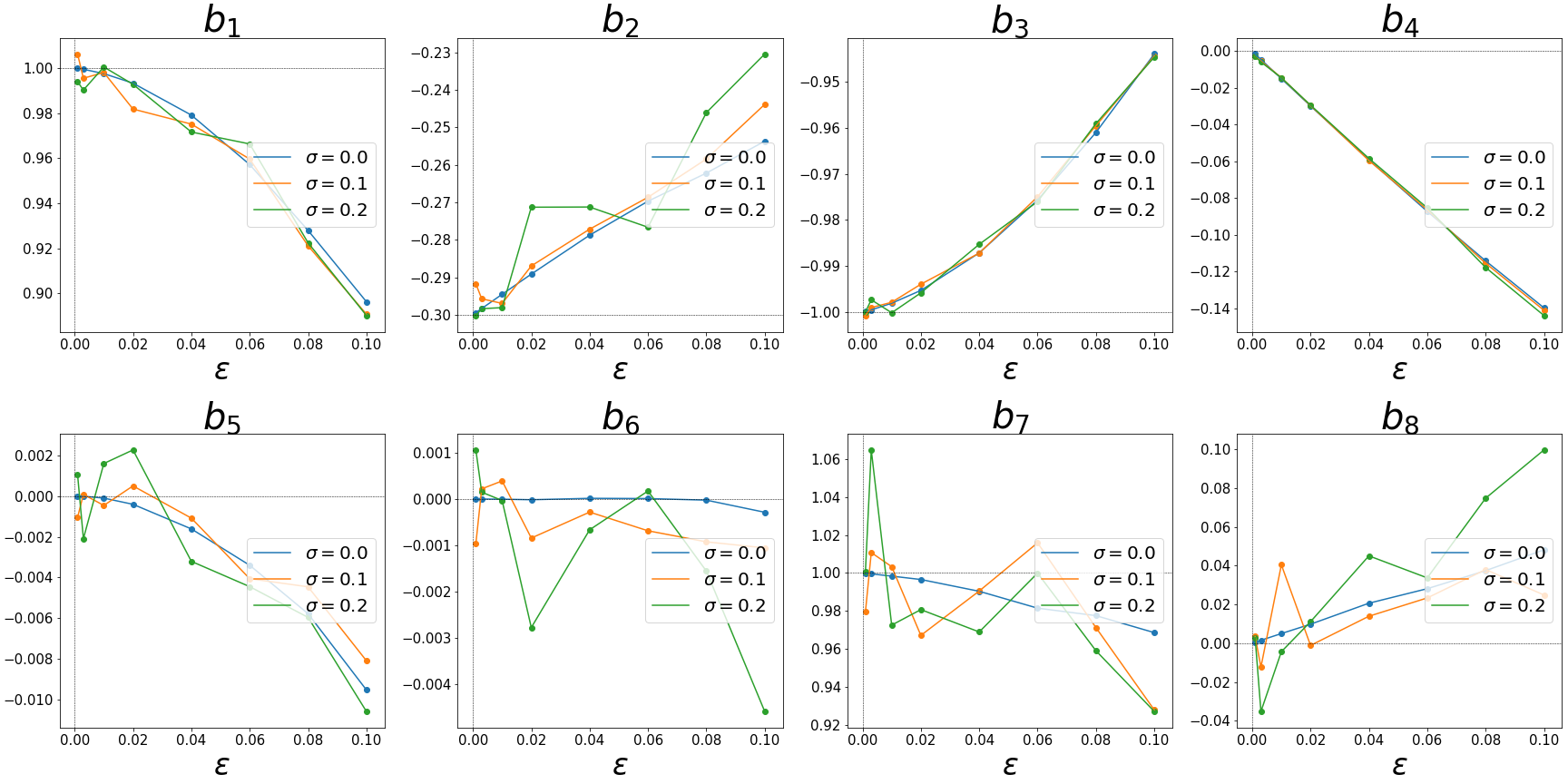}
\caption{Plots of the $\epsilon$-dependence of the $b_i$ coefficients ($i=1,...,8$) in the model $h(X)$ (cf. Eq.~\ref{eqn:Duffing_mapping}).
The intersection of the thin vertical and horizontal gray lines indicate the correct value at $\epsilon = 0$.}
\label{fig:Duffing_DE_params_b}
\end{figure*}

The process of fitting $Y$ to $X$ for a single $\epsilon$ is now repeated for the set of $\epsilon$-values  $\{0.001, ...0.1\}$,
leading to the plots in Figs.~\ref{fig:Duffing_DE_params_a},~\ref{fig:Duffing_DE_params_b}; these plots include data for the cases $\sigma=0, 0.1, \text{and } 0.2$.
Best line fits were used to model the $\epsilon$-dependence, with the results collected in Table~\ref{tab:fits} in Appendix~\ref{sec:app-params};
in three of the cases it was appropriate to use polynomial regression up to second-order.
These fits were extrapolated to $\epsilon = 0$ to determine parameter values for the underlying DE.
Equation~\ref{eqn:Duffing_mapping} may now be rewritten in this limiting case, and the variables will be written using the standard differential notation 
(i.e., $\epsilon \rightarrow dt$, $\Delta u \rightarrow du$, $\Delta \dot{u} \rightarrow d\dot{u}$ ), producing
\begin{subequations}
\begin{align}
du & = dt \, [ \dot{u} ] \label{eqn:Duffing_first} \\
d\dot{u} & = dt \, [  -0.3 \dot{u} + u - u^3 + p]  \, . \label{eqn:Duffing_second}
\end{align}
\end{subequations}
The first of these equations (Eq.~\ref{eqn:Duffing_first}) is just a contact condition, and reflects the fact that in the space $(u,\dot{u})$, not all three
quantities $du$, $dt$, $\dot{u}$ are independent (i.e., $\dot{u} = du/dt$).
The combination of these two equations is just equivalent to the original equation of motion for the Duffing oscillator (Eq.~\ref{eqn:duffing_both}),
with the chosen parameter values.

\subsection*{Extrapolation}

Using $\epsilon = 0.1$ and the values for $a_i$ and $b_i$ found earlier, an extrapolation of $u(t)$
can be created using an extension of the earlier generative algorithm, here shown as Algo.~\ref{algo:generative_Duffing}.
\begin{algorithm}[H]
\caption{: Generation step for Duffing oscillator}
\begin{algorithmic}
\STATE{ Input: $t_0$, $N$, $\{p\}$, $\{\dot{p}\}$ }
\STATE{ Init: $t = t_0$ }
\STATE{ Init: $u = u(t_0)$ }
\STATE{ Init: $\dot{u} = \dot{u}(t_0)$ }
\FOR{ i = $0$ to $N$ }
\STATE{ $	\text{tmp\_}h_1      = h_1(u,\dot{u}, p(t), \dot{p}(t))  $ }
\STATE{ $	\text{tmp\_}h_2 = h_2(u,\dot{u}, p(t), \dot{p}(t))  $ }
\STATE{ $	u \leftarrow u + \text{tmp\_}h_1 $ }
\STATE{ $	\dot{u} \leftarrow \dot{u} + \text{tmp\_}h_2 $ }
\STATE{ $t \leftarrow t + \epsilon$ }
\STATE{ print($u,t$) }
\ENDFOR
\end{algorithmic}
\label{algo:generative_Duffing}
\end{algorithm}
This extension is made because during training the values of $p$ and $\dot{p}$ are recorded (as scalars), and are included in the feature set $X$ for training.  
Thus, during extrapolation, new values for $p$ and $\dot{p}$ are fed into this algorithm.  
(The notation $\{ p \}, \{ \dot{p} \}$ is used to denote that {\em functions} for $p$,$\dot{p}$ are being passed in to this algorithm, and are evaluated at time $t$.)
In principle any smooth function can be used for $p(t)$.

\begin{figure}
\includegraphics[scale=0.27]{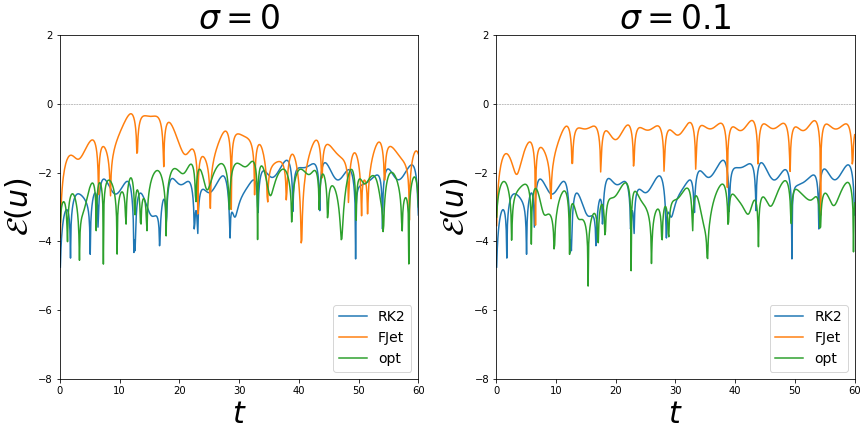}
\caption{
Shown are plots of the errors for the Duffing oscillator.
Here the {\em opt} solution and RK2 are comparable.  In this case the FJet solution fares slightly worse.}
\label{fig:Duffing_extrap}
\end{figure}

Shown in Fig.~\ref{fig:Duffing_extrap} are plots of the error with respect to RK4 (using Eq.~\ref{eqn:param-error}) for the cases $\sigma=0$ and $0.1$, with initial condition $(u,\dot{u}) = (1,0)$.
The data are from RK2, FJet, and the optimized FJet.
In this case, FJet performs comparably to RK2 when there is no noise, but slightly worse when noise is present.
In both cases, the optimized version of FJet performs at least as well as RK2, in general.


\section{Function Space for $h(X)$}
\label{sec:fcnspace}

In Step \#8 of Sec.~\ref{sec:methodology}, it was explained how the feature set $X_n$ could be derived from a numerical integration scheme
by Taylor series expanding it, and then collecting the functions that contribute at ${\cal O} (\epsilon^n)$.
(For simplicity, the same feature set will be used for $\Delta u$ and $\Delta \dot{u}$.)
This was done for all the examples in Appendix \ref{sec:app-RK}; for convenience, the results for the pendulum are collected here in Table~\ref{tab:fcn-space}.
\begin{table}[b]
\caption{The required feature set $X_n$ for the pendulum example to achieve accuracy ${\cal O}(\epsilon^n)$ in an RK context.  
It is shown for $n=2$ (Euler), $n=3$ (RK2), and $n=5$ (RK4).
}
\begin{ruledtabular}
\begin{tabular}{ll} 
  $n$ &
  $X_n$ \\
\colrule
$2$ &  $\{ \dot{u}, \sin u \}$  \\ 
$3$ &  $\{ \dot{u}, \sin u, \dot{u}\cos u \}$  \\ 
$5$  & $\left\{ \dot{u}, \sin u, \dot{u}\cos u, \sin u \cos u, \right. $ \\
& $\left. \dot{u} \sin^2 u, \dot{u}^2 \sin u, \dot{u}^3 \cos u  \right\} $
\end{tabular}
\end{ruledtabular}
\label{tab:fcn-space}
\end{table}
The main point of this is that when using feature regression on these zero-noise examples, it's possible to determine
a set of features which lead to an error estimate $\sim {\cal O}(\epsilon^n)$.
Indeed, these functions ($X_n$) define a function space \cite{Munkres-2015,Wasserman_functionspaces} for a mapping from $(u,v)$ to $Y=(\Delta u,\Delta v)^T$, 
and yield a convergence criterion of
\begin{align}
Y - h(X_n) \sim {\cal O} (\epsilon^n)
\label{eqn:convergence}
\end{align}
where again, it is assumed the data are noiseless.
Thus, in this case the (feature regression) model $h$ converges to $Y$ as $n \rightarrow \infty$ and/or $\epsilon \rightarrow 0$.  (One can also set $\epsilon =1/n$.)
An important point to recognize though is that the codomain $Y$ is actually {\em constrained}: it is required to represent the incremental updates 
corresponding to the dynamics/DE  being modeled (i.e., it's an estimate of the tangent space using time step $\epsilon$).  There is no constraint for the domain $(u,v)$.
Also, even though only $2$nd-order systems are being discussed here, higher order ones may also be used.

More generally, these criteria (the domain, codomain, convergence criterion) may be used as a {\em definition} for the function space of a class of functions that model
data which originated from dynamics governed by the appropriate DE.  
To that end, besides using feature regression, other models could be used, such as a NN.
Also, it might be worthwhile to use the $X_n$ as input features in the NN.

The reader might also note that the sets $X_n$ are nested, i.e. $X_n \subseteq X_{n+1}$.
Also, the disjoint sets $A_2 = X_2$, $A_3 = X_3 - X_2$, $A_4 = X_4 - X_3$, ... comprise a graded function space, 
and the set $X_n = \bigoplus_{i=2}^n A_i$ can be used to achieve ${\cal O}(\epsilon^n)$ accuracy.

In comparison to the function space for a Fourier series, the constraint on $Y$ appears much stronger:
with Fourier series it is only required that the function being modeled be periodic, piecewise $C^1$.
Also, one might note that in Eq.~\ref{eqn:convergence}, 
$h(X_n)$ supplants the role of a partial sum of the Fourier series \citep{Olver-PDE-2020} in a similar convergence criterion.

Finally, the reader should understand that this section contains two new results for this class of regression-type approaches.
It is the first time that (1) the uncertainty has been quantified for a dynamical model derived from ML;
(2) the set of features ($X_n$) has been derived in a principled way.


\section{Regularization}
\label{sec:regu}

In previous work starting with \cite{Crutchfield-1987}, the norm has been to use a single, small value of $\epsilon$ along with a regularization term, and from that derive the underlying DE.  In contrast, no explicit regularization term was used in this work, and instead the $\epsilon$-dependence was determined for each parameter of the candidate features.  Then, by extrapolating these $\epsilon$-dependencies to $\epsilon =0$, the underlying DE was determined.

However, any limiting of the complexity of a model is a form of regularization, and thus it was present implicitly in this paper.
In particular, it was present due to the set of initial features being {\em finite}.  It was also present when this set of features was iteratively adjusted (cf. ch. 3 in \cite{Hastie_ESL2}) during the fitting process.  
Thus, in this paper, determining the features was more of a iterative process (involving implicit regularization) rather than a direct computation.



\section{Comparisons}
\label{sec:comp}

\begin{figure}
\includegraphics[scale=0.27]{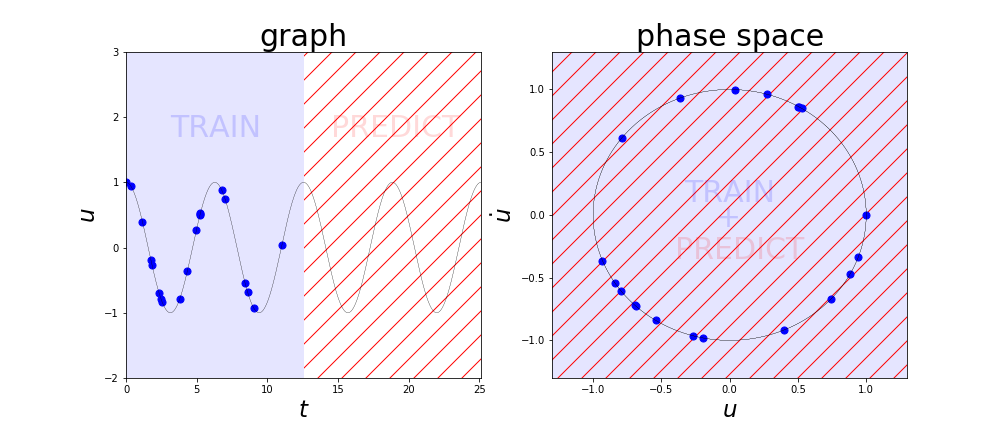}
\caption{
Training is done in the regions shaded blue, while prediction (extrapolation) is done in the regions shaded with red lines.
The standard approach involves modeling a system using the graph $(t,u)$, where the training/prediction regions are disjoint.
This is a problem for ML since it generally can only interpolate.
In contrast, in the phase space approach (in right plot), these regions are the same, and hence there are no issues.
}
\label{fig:graph_phase}
\end{figure}

{\em \underline{Phase Space Approach} }:
At the outset of this paper, the motivation had been to find a way to solve the ``extrapolation problem" of modeling dynamical systems with ML:
that is, the problem of training a model (which interpolates) in one domain and trying to use it in another domain.
Thus, if one models the system's behavior (i.e., $u(t)$) over some finite domain of times, one has a model {\em for that set of times};
using that model for later times violates common wisdom about how to apply ML models.  
(This has been the standard approach for over 30 years \cite{Crutchfield-1987}, and has been used recently in \citep{Rudy-2017}, for example.)
In contrast, the FJet approach uses ML to model the updates in phase space.
This means that the training and subsequent prediction/extrapolation are done over the same domain, so interpolation is a non-issue.
This difference is illustrated in Fig.~\ref{fig:graph_phase},
which shows the domains related to training and prediction for the ``standard" approach (cf. left plot),
and the phase space approach (cf. right plot).
Thus, the extrapolation problem can be circumvented by recollecting that dynamics can be equally well described \cite{Atlee-book-v1} in phase space $(u,\dot{u})$ as in a graph $(t,u)$.

Finally, there were several interesting papers \cite{Gouesbet-1991,Gouesbet-1992} where the authors also used ML to study the dynamics in the phase space of similar systems.
However, outside of these commonalities, their approach was very different.

{\em \underline{Function Space}}:
One aspect of the standard approach is that the modeling is done using a set of functions that is poorly motivated.
It has been advocated for as a sort of expansion, but it is never stated what exactly is being expanded.
Again, this has been done since the standard approach was introduced \cite{Crutchfield-1987} and it continues to recent times \citep{Rudy-2017,Aguirre-review-2009}.

In contrast, in this paper a principled derivation of the function space was given.  
It was shown that a specific set of functions should be used in order to achieve a specified level of uncertainty quantification.
As this was presented, it may appear to the reader to only be a statement of the {\em existence} of such a function space given the DE.
However, as discussed in Sec.~\ref{sec:fcnspace}, one can use a trial set of functions, compute the underlying DE that follows from them, and then go on
to expand the RK implementation of that DE to determine the associated function space.  
A self-consistency check requires that the initial set of functions and the functions derived from the DE be the same.

{\em \underline{Uncertainty Quantification}}:
This paper demonstrated how to achieve uncertainty quantification using FJet and feature regression.
Doing so has long been a goal with this kind of modeling, but to the author's knowledge, has never been achieved elsewhere with such ML models.

{\em \underline{Underlying DE}}:
To the author's knowledge, in all other papers where the underlying DE was derived, only a single (small) value for the time step ($\epsilon$) would be used.
However, as shown by the author (e.g., Fig.~\ref{fig:HO_DE_params}), values at a single $\epsilon$ can have a wide variance, and are less useful in determining the underlying DE.
In contrast, in this paper, a number of such small $\epsilon$ values 
were used (cf. Figs.~\ref{fig:HO_DE_params},\ref{fig:Pendulum_DE_params},\ref{fig:Duffing_DE_params_a},\ref{fig:Duffing_DE_params_b}), 
in order to provide for a more robust fit than that afforded by a single $\epsilon$.
The fit was done with a low-order polynomial (mostly linear) and extrapolated to $\epsilon=0$.
This is appropriate for determining the DE, as it involves infinitesimal changes in time.

\section{Final Remarks}
\label{sec:final}

What has been achieved in this paper are the following:
\begin{itemize}
\item Introduced FJet: new way of employing ML to model dynamical systems
\item Derived tangent space for solution manifold using FJet
\item Achieved uncertainty quantification of solution in terms of ${\cal O}(\epsilon^n)$
\item Introduced principled derivation of a function space for modeling
\item Related a graded function space to the order of uncertainty quantification
\item Demonstrated FJet on several examples, including one with time-dependent forcing
\end{itemize}

The FJet method has been introduced as an alternative means to model dynamical data and produce the underlying DE.
It does so by performing ML modeling on updates in the phase space of the system.
More generally the space is $U^{(n-1)}$, when the underlying equation is of order $n$.
In the FJet approach, the update data is formed by storing differences in the phase space variables over a small time step.
In contrast, competing methods typically perform modeling over the time domain (cf. Sec.~\ref{sec:background}).


The FJet approach can be understood from the vantage point of the numerical integration scheme Runge-Kutta.
By expanding an RK scheme with respect to the time step ($\epsilon$), certain features automatically appear.  
These features can be identified as the appropriate functions to use in a feature regression for a given example;
this was demonstrated for all three examples.
In particular, in the Duffing oscillator example (cf. Eq.~\ref{eqn:Duffing_mapping}), 
the identified features included terms for an external time-dependent force, in analogy to a Green's function approach.  
Finally, these considerations led to a general recommendation (Eq.~\ref{eqn:rubric}) for what 
features to include in the normal experimental scenario where the equation of motion isn't known beforehand.
Also, understanding the origin of the features in this way is instructive in other ways.
For example, if one wished to study the Duffing oscillator with {\em multiplicative} coupling of the forcing term $p(t)$ with $u$, then the approach here would indicate how to modify the feature set.


While the modeling in this paper was done exclusively using feature regression, one should not make the mistake of thinking that it is a requirement.  
As pointed out at several points in the paper, any regression technique may be used.  
The reason for using feature regression is that it facilitates determining the underlying DE. 
If that is not a goal of the researcher, then other ML methods (such as NNs, XG-Boost, Random Forests) become competitive options.


In the three examples considered, it was demonstrated how the FJet model could lead to extrapolations beyond the domain of the original training data.  
The accuracy of this extrapolation depends upon the number of variables, as well as the quality of the initial data for the updates (i.e., $\Delta u$, $\Delta \dot{u}$).  
In general, it was seen that if the feature set $X$ included all the features from RK2 (after it was Taylor series expanded, as in Appendix~\ref{sec:app-RK}), then it could be expected to be as accurate as RK2.  The same can be said with respect to RK4.
However, in the case where $X$ happens to be the {\em exact} feature set, as may happen in linear problems, it becomes possible 
to obtain an extremely accurate solution, one that is accurate up to the machine precision where the computations are done.  
Such was the case with the harmonic oscillator example.


The initial features that are considered herein are relatively complex (e.g., $ \dot{u} \sin u$ in the Pendulum example) compared to other models, which commonly use a simple polynomial in $u$, for example.
It is here asserted that there is an interplay between the complexity of the initial features, and the complexity of the remaining ML model.  
In particular, it is asserted that for models of comparable (predictive) quality, the ``sum" of these two complexities is approximately constant.  
The author advocates for models with relatively complex initial features (as they make it easier to train and understand the model), whether the remaining model be relatively simple (regression) or complex (NNs).  
Also, the initial features are a good place for the researcher to incorporate any subject matter expertise (e.g., constraints, symmetries, physical laws, or heuristics).
Finally, it was seen in the examples that such complex initial features simplify the determination of the underlying DE.

It's also noted that that for linear DEs, there is at some level a similarity between FJet and the much older method of solving DEs using {\em isoclines} \citep{Rainville-1983}.
Recall that with isoclines, one is given $du/dt = f(u)$ (for some function $f$), and then draws the magnitude/direction of an update vector
throughout the $(u,t)$ plane.  Following that, one draws the updates (forming a curve) that are consistent with these local tangents, and so obtains an approximate solution.
The analogy to FJet is that at each point either method uses local information to decide where to update the solution: for isoclines one has local tangents,
while for FJet one has the machine-learned models $h_1$, $h_2$.  Aside from this similarity, there are of course many differences.

Finally, a very high-level comparison can be made between FJet and Zero-Shot Learning (ZSL) \cite{Larochelle-2008},
which is a classification technique in ML used to make predictions on images that are different from the types it was trained on.
In a sense, they both {\em extrapolate} beyond the domain of their training data (i.e., their in-sample data):
FJet extrapolates beyond the domain of in-sample times; ZSL extrapolates beyond the set of in-sample images.
In both cases, these methods succeed by exploiting something that connects the in-sample and (unrepresented) out-of-sample data:
for FJet, it's a common dynamical model; for ZSL, it's a common textual description.

\section*{Acknowledgement}

The author found general inspiration for considering this subject from attending the Kavli Institute for Theoretical Physics conference {\em At the Crossroads of Physics and Machine Learning}, Santa Barbara, Feb 2019.
Also, the author kindly acknowledges the insightful instruction on differential equations and dynamical systems from Prof. Julian I. Palmore and Prof. E. Atlee Jackson (d.) at U. Illinois at Urbana-Champaign during the author's studies there.

\appendix

\section{Runge-Kutta}
\label{sec:app-RK}

The Runge-Kutta scheme \citep{Runge-1895,Kutta-1901,Heun-1900,Hairer-1987} is a standard technique for the numerical integration of a solution
given its ODE.  An $n$th-order explicit ODE
can be rewritten as a system of first-order DEs, and thus the general form for the ODE can be expressed as
\begin{align}
\frac{dy}{dt} = f(t,y)
\end{align}
where $t (y)$ represents the independent (dependent) variables.
The RK scheme involves an update from $y_n$ to $y_{n+1}$, where the difference is
\begin{align}
\Delta y_n & \equiv y_{n+1} - y_n \, .
\end{align}
After defining these quantities
\begin{align}
k_1 & = \epsilon f(t_n, y_n) \\
k_2 & = \epsilon f(t_n + \frac{1}{2} \epsilon, y_n + \frac{1}{2} k_1) \\
k_3 & = \epsilon f(t_n + \frac{1}{2} \epsilon, y_n + \frac{1}{2} k_2) \\
k_4 & = \epsilon f(t_n + \epsilon, y_n + k_3) 
\end{align}
the schemes for Euler, RK2 and RK4 can be summarized in Table \ref{tab:RK}.
\begin{table}[b]
\caption{Update values $\Delta y_n$ for three numerical integration schemes.  Note the different levels of accuracy in the time step $\epsilon$.}
\begin{ruledtabular}
\begin{tabular}{lcr} 
\textbf{Scheme}  &  $\Delta y_n$  & \textbf{Accuracy}  \\
\colrule
  Euler  &     $k_1$    &   $O(\epsilon^2)$   \\
  RK2   &      $k_2$    &   $O(\epsilon^3)$   \\
  RK4   &      $\frac{1}{6} (k_1 + 2 k_2 + 2 k_3 + k_4)$    &   $O(\epsilon^5)$   
\end{tabular}
\end{ruledtabular}
\label{tab:RK}
\end{table}
Note how complexity is built up in this scheme.  For nonlinear ODEs this can lead to many terms involving powers and derivatives of the original
variables.

In the subsections to follow, $\Delta y$ will be computed in the RK2 scheme and then expanded with respect to $\epsilon$;
importantly, this expansion is done to the same order in $\epsilon$ as the original RK expression; recall that 
RK is itself motivated from a Taylor expansion of a solution (cf. Ch.2 of \cite{Hairer-1987}, or App.B of \citep{Ince-1956}).
Since FJet are models of $\Delta y$, this expansion essentially becomes a derivation of the models $h=(h_1,h_2, \dots)$
in the feature regression technique.
In the examples to follow, $y = (u,\dot{u})$, and so
\begin{align*}
\Delta y  =
\begin{bmatrix}
\Delta u \\
\Delta \dot{u}
\end{bmatrix}
 =
\begin{bmatrix}
h_1 \\
h_2
\end{bmatrix}
\end{align*}
where $h_1$ and $h_2$ are functions of $(u,\dot{u})$.
Also, throughout this appendix, the substitution $v = \dot{u}$ will be used for the RK expansions.

\subsection*{Harmonic Oscillator}

The equation for the damped harmonic oscillator (Eq.~\ref{eqn:HO}) can be written as
\begin{align*}
\begin{bmatrix}
\dot{u}\\
\dot{v}
\end{bmatrix}
& =
\begin{bmatrix}
v\\
 - \omega_0^2 u -2\gamma v
\end{bmatrix} 
=  f
\left(
\begin{bmatrix}
u\\
v
\end{bmatrix}
\right)  \, .
\end{align*}
Note that there is no explicit time dependence in $f$.
The RK2 scheme produces
\begin{subequations}
\begin{align}
\begin{bmatrix}
h_1 \\
h_2
\end{bmatrix}
& =  \epsilon 
\begin{bmatrix}
v  - \frac{\epsilon}{2} \omega^2  u -\epsilon \gamma v \\
 -\omega_0^2  ( u + \frac{\epsilon}{2} v )  -2\gamma (v - \frac{\epsilon}{2} \omega_0^2 u -\epsilon \gamma v)
\end{bmatrix} \nonumber \\
& =  \epsilon 
\begin{bmatrix}
- \frac{\epsilon}{2} \omega_0^2  &  1 - \epsilon \gamma \\
-\omega_0^2 (1 - \epsilon \gamma)  &   -2\gamma (1- \epsilon \gamma ) -\frac{\epsilon}{2} \omega_0^2
\end{bmatrix}
\begin{bmatrix}
u \\
v
\end{bmatrix}  \, . 
\label{eqn:h-RK2-HO}
\end{align}
\end{subequations}
This shows how the vector $(u,v )^T$ naturally appears as the features.
Since this is a linear differential equation, the same set of features results for RK4.

\subsection*{Pendulum}

Equation~\ref{eqn:pendulum} can be rewritten as
\begin{align*}
\begin{bmatrix}
\dot{u}\\
\dot{v}
\end{bmatrix}
& =
\begin{bmatrix}
v\\
 -2\gamma \dot{u} -\omega_0^2\sin u
\end{bmatrix}
 = f
\left(
\begin{bmatrix}
u\\
v
\end{bmatrix}
\right)  \, . 
\end{align*}
Note that there is no explicit time dependence in $f$.
The RK2 scheme produces Eq.~\ref{eqn:h-RK2-pendulum}
\begin{widetext}
\begin{subequations}
\begin{align}
\begin{bmatrix}
h_1 \\
h_2
\end{bmatrix} 
& =  \epsilon
\begin{bmatrix}
(1 - \epsilon \gamma) v - \frac{\epsilon}{2} \omega_0^2 \sin u \\
-2\gamma (1 - \epsilon \gamma) v + \epsilon \gamma \omega_0^2 \sin u - \omega_0^2 \sin (u + \frac{\epsilon}{2} v)
\end{bmatrix} \nonumber \\
& =  \epsilon 
\begin{bmatrix}
(1 - \epsilon \gamma )  &  - \frac{\epsilon}{2} \omega_0^2  &  0  \\
-2\gamma (1 - \epsilon\gamma ) &   -\omega_0^2 (1 - \epsilon\gamma ) &  -\frac{\epsilon}{2} \omega_0^2  
\end{bmatrix}
\begin{bmatrix}
v \\
\sin u \\
v \cos u 
\end{bmatrix}
 + O(\epsilon^3)  \, . 
\label{eqn:h-RK2-pendulum}
\end{align}
\end{subequations}
\end{widetext}
Observe that since this is being compared to RK2, a difference of ${\cal O}(\epsilon^3)$ in Eq.~\ref{eqn:h-RK2-pendulum} is irrelevant.

Thus, this expansion in $\epsilon$ of RK2 produces a model akin to the feature regression technique,
with the set of predictor variables being $\{ v, \sin u, v\cos u \}$.
If instead, the RK4 scheme were expanded in $\epsilon$, the features for $\Delta u$ would be
$\{ v, \sin u, v \cos u, \sin u \cos u, v^2 \sin u \}$, while for $\Delta \dot{u}$ they would additionally include
$\{ v \sin^2 u, v \cos^2 u, v^3 \cos u \}$.
The contribution of $v \cos^2 u$ can be absorbed into $\{ v, v\sin^2 u \}$; it would otherwise lead to a collinearity during regression.
Thus, RK4 produces the additional features $\{ \sin u \cos u, v \sin^2 u, v^2 \sin u, v^3 \cos u \}$, compared to that for RK2.
Also, these nonlinear features remain in the set even if $\gamma=0$.

\subsection*{Duffing Oscillator}


Equation~\ref{eqn:duffing_both} can be rewritten as
\begin{align*}
\begin{bmatrix}
\dot{u}\\
\dot{v}
\end{bmatrix}
& = f
\left( t,
\begin{bmatrix}
u\\
v
\end{bmatrix}
\right)
=
\begin{bmatrix}
v\\
-2\gamma v - \alpha u -\beta u^3 + p(t)
\end{bmatrix} \, . 
\end{align*}
Note that in this case there is explicit time dependence in $f$ due to $p(t)$.
The RK2 scheme produces Eq.~\ref{eqn:h-RK2-Duffing}
\begin{widetext}
\begin{subequations}
\begin{align}
\begin{bmatrix}
h_1 \\
h_2 
\end{bmatrix}
& =  \epsilon 
\begin{bmatrix}
(1 - \epsilon \gamma ) v - \frac{\epsilon}{2} \alpha u - \frac{\epsilon}{2} \beta u^3 + \frac{\epsilon}{2} p(t) \\
-2\gamma (1 - \epsilon \gamma) v  - \frac{\epsilon}{2} \alpha v - \alpha (1 - \epsilon\gamma )u  
+ \epsilon \beta \gamma u^3  -\beta ( u + \frac{\epsilon}{2} v )^3 
  - \epsilon \gamma p(t) + p(t + \frac{\epsilon}{2})
\end{bmatrix} \nonumber \\
& =  \epsilon 
\begin{bmatrix}
- \frac{\epsilon}{2} \alpha  & (1 - \epsilon \gamma )  &  -\frac{\epsilon}{2} \beta  &  0  &  \frac{\epsilon}{2}  &  0 \\
 -\alpha (1 - \epsilon \gamma ) & -2\gamma (1 - \epsilon \gamma ) -\frac{\epsilon}{2} \alpha &   -\beta (1 - \epsilon \gamma)
 & -\frac{3\epsilon}{2} \beta  & (1 - \epsilon \gamma) & \frac{\epsilon}{2} 
\end{bmatrix}
\begin{bmatrix}
u \\
v \\
u^3 \\
u^2 v \\
p \\
\dot{p}
\end{bmatrix}
 + O(\epsilon^3) 
\label{eqn:h-RK2-Duffing}
\end{align}
\end{subequations}
\end{widetext}
where $\dot{p}$ denotes the time derivative of $p$.
Once again, the ${\cal O}(\epsilon^3)$ expansion is an equivalent rewriting, since this is for RK2.
Thus, this numerical integration scheme implies the vector of predictor variables $X = ( v, u, u^3, u^2 v, p, \dot{p})^T$
if feature regression is used.

\subsection*{General Expansion}

This section is meant to provide some guidance for choosing an initial set of features in the FJet approach.
It is motivated by a Taylor series expansion of the RK2 scheme for $\frac{dy}{dt} = F(t,y)$:
\begin{align*}
\Delta y 
& =  \epsilon F + \frac{1}{2} \epsilon^2 \partial_t F + \frac{1}{2} \epsilon^2 F \partial_y F + O(\epsilon^3) \, . 
\end{align*}
Upon setting $F=(f,g)^T$, this appears as
\begin{align*}
\begin{bmatrix}
\Delta u \\
\Delta \dot{u}
\end{bmatrix}  
& =
\epsilon 
\begin{bmatrix}
f \\
g 
\end{bmatrix}  
+
\frac{\epsilon^2}{2}
\begin{bmatrix}
f_t \\
g_t 
\end{bmatrix}  
+
\frac{\epsilon^2}{2}
\begin{bmatrix}
f_u & f_{\dot{u}} \\
g_u & g_{\dot{u}}
\end{bmatrix}  
\begin{bmatrix}
f \\
g 
\end{bmatrix}  
+ O(\epsilon^3) \, . 
\end{align*}
This expansion explicitly produces the monomials of $f$ and $g$ that are needed.
Thus, if in the beginning of modeling, one only suspected that $f$ and $g$ may be features for $\Delta u$ and $\Delta \dot{u}$, 
it would follow that the features $\{ f, f_t, ff_u, gf_{\dot{u}} \}$ should be used for $\Delta u$, 
and $\{ g, g_t, fg_u, gg_{\dot{u}} \}$ should be used for $\Delta \dot{u}$.
Continuing on with this logic, one might pursue an expansion of the RK4 scheme for even higher accuracy.
However, the complexity becomes significant enough to make it less useful.
(Again, the reader is reminded that RK is itself based on a Taylor series expansion, but RK is preferred here since it's already being used in this paper.)

Instead, the approach will be used of more easily generating a superset of features, which will include those found by expanding RK2 or RK4.
From that superset, the unnecessary additional terms can be eliminated during the fitting process of FJet.
Indeed, if ${\cal F}$ is the original set of features, the superset consists of the terms appearing in
\begin{equation}
\sum_{n=1} ( {\cal F} + \partial_J {\cal F} + \partial_J \partial_J {\cal F} + \cdots )^n
\label{eqn:rubric}
\end{equation}
where $J$ denotes any of the jet space variables being used, and $\partial_J$ includes all possible partial derivatives with respect to $J$.
For example, in the harmonic oscillator and pendulum examples, $J = \{ u, \dot{u} \}$; for the Duffing oscillator $J = \{ t, u, \dot{u} \}$.
The time derivative would only be applied to variables with an explicit time dependence, such as $p(t)$ and not $u$ or $\dot{u}$.
Also, if there are any special symmetry considerations, that should also be accounted for during feature selection.
Finally, the extent to which every possible monomial actually contributes depends on the amount of nonlinearity in the underlying DE.  
In particular, linear problems like the harmonic oscillator have no such nonlinear or derivative contributions.

\section{Parameter Fitting}
\label{sec:app-params}

As shown in Table~\ref{tab:fits}, the linear extrapolations between RK2 and FJet (for $\sigma=0,0.1$) agree well.
\begin{table}[b] 
\caption{\label{tab:fits}%
Linear fits for each parameter as a function of the time step $\epsilon$.
The three sections, from top to bottom, correspond to the harmonic oscillator, the pendulum, and the Duffing oscillator.
The values expected from RK2 (cf. Appendix~\ref{sec:app-RK}) are in the second column; the third and fourth columns contain the fits from FJet for the cases
of $\sigma = 0$ and $\sigma = 0.2$.
A quadratic fit in $\epsilon$ was used for $b_1$, $b_3$, and $b_5$ in the Duffing oscillator example.
}
\begin{ruledtabular}
\begin{tabular}{lrrr}
  \textrm{}  &
  \textrm{RK2}  &
  \textrm{FJet}$_0$    &
  \textrm{FJet}$_{0.2}$  \\
\colrule
$a_1$ & $0-0.5\epsilon$ & $-0.000 -0.496 \epsilon$ & $-0.001 -0.507 \epsilon$ \\
$a_2$ & $1-0.1\epsilon$ & $1.000 -0.115 \epsilon$ & $0.998 -0.134 \epsilon$ \\
$b_1$ & $-1 + 0.1\epsilon$ & $-1.000  + 0.115 \epsilon$ & $-1.001  + 0.092 \epsilon$ \\
$b_2$ & $-0.2-0.48\epsilon$ & $-0.200 -0.473 \epsilon$ & $-0.198 -0.555 \epsilon$ \\
\colrule
$a_1$ & $1-0.1\epsilon$ & $1.000 -0.099 \epsilon$ & $0.999 -0.109 \epsilon$ \\
$a_2$ & $0-0.5\epsilon$ & $-0.000 -0.496 \epsilon$ & $0.001 -0.509 \epsilon$ \\
$a_3$ & $0 + 0\epsilon$ & $0.000 -0.016 \epsilon$ & $0.002 -0.077 \epsilon$ \\
$b_1$ & $-0.2 + 0.02\epsilon$ & $-0.200  + 0.019 \epsilon$ & $-0.199 -0.022 \epsilon$ \\
$b_2$ & $-1 + 0.1\epsilon$ & $-1.001  + 0.152 \epsilon$ & $-1.003  + 0.129 \epsilon$ \\
$b_3$ & $0-0.5\epsilon$ & $-0.000 -0.492 \epsilon$ & $0.001 -0.490 \epsilon$ \\
\colrule
$a_1$ & $0 + 0.5\epsilon$ & $0.000  + 0.475 \epsilon$ & $0.005  + 0.454 \epsilon$ \\
$a_2$ & $1-0.15\epsilon$ & $1.000 -0.134 \epsilon$ & $0.994 -0.129 \epsilon$ \\
$a_3$ & $0-0.5\epsilon$ & $-0.000 -0.486 \epsilon$ & $-0.001 -0.483 \epsilon$ \\
$a_7$ & $0 + 0.5\epsilon$ & $0.000  + 0.491 \epsilon$ & $0.008  + 0.453 \epsilon$ \\
$b_1$ & $1-0.15\epsilon$ & $1.000 -0.224 \epsilon$ & $0.994  + 0.070 \epsilon$ \\
$b_2$ & $-0.3 + 0.545\epsilon$ & $-0.299  + 0.465 \epsilon$ & $-0.299  + 0.642 \epsilon$ \\
$b_3$ & $-1 + 0.15\epsilon$ & $-1.000  + 0.172 \epsilon$ & $-1.000  + 0.202 \epsilon$ \\
$b_4$ & $0-1.5\epsilon$ & $-0.001 -1.407 \epsilon$ & $-0.001 -1.435 \epsilon$ \\
$b_5$ & $0 + 0\epsilon$ & $-0.000  + 0.001 \epsilon$ & $0.001 -0.015 \epsilon$ \\
$b_6$ & $0 + 0\epsilon$ & $0.000 -0.002 \epsilon$ & $0.000 -0.034 \epsilon$ \\
$b_7$ & $1-0.15\epsilon$ & $1.001 -0.314 \epsilon$ & $1.014 -0.748 \epsilon$ \\
$b_8$ & $0 + 0.5\epsilon$ & $0.000  + 0.475 \epsilon$ & $-0.015  + 1.117 \epsilon$ \\
\end{tabular}
\end{ruledtabular}
\end{table}
The differences between the values at $\epsilon=0$ lead to the errors in Table~\ref{tab:errors}.
The errors when $\sigma=0$ are small, but as expected they increase slightly for increasing $\sigma$.
\begin{table}[b] 
\caption{\label{tab:errors}%
Errors for the $\epsilon=0$ extrapolation of the FJet model of the harmonic oscillator for the cases of $\sigma = 0,0.1, 0.2$.
The error is measured by Eq.~\ref{eqn:param-error}.
The three sections, from top to bottom, correspond to the harmonic oscillator, the pendulum, and the Duffing oscillator.
As expected, the errors decrease for smaller noise levels ($\sigma$).
}
\begin{ruledtabular}
\begin{tabular}{lccr}
  \textrm{}&
  ${\cal E}_0$ &
  ${\cal E}_{0.1}$ &
  ${\cal E}_{0.2}$ \\
\colrule
$a_1$ & -4.43 & -3.58 & -3.02 \\ 
$a_2$ & -3.81 & -3.25 & -2.77 \\ 
$b_1$ & -3.81 & -2.86 & -2.91 \\ 
$b_2$ & -4.17 & -3.23 & -2.75 \\ 
\colrule
$a_1$ & -5.19 & -3.34 & -2.90 \\ 
$a_2$ & -4.31 & -3.54 & -2.91 \\ 
$a_3$ & -3.79 & -2.81 & -2.73 \\ 
$b_1$ & -5.37 & -3.56 & -2.97 \\ 
$b_2$ & -3.27 & -2.46 & -2.54 \\ 
$b_3$ & -4.05 & -2.98 & -3.22 \\
\colrule
$a_1$ & -3.50 & -2.80 & -2.31 \\ 
$a_2$ & -3.81 & -3.33 & -2.25 \\ 
$a_3$ & -3.76 & -4.04 & -3.05 \\ 
$a_7$ & -4.01 & -3.41 & -2.11 \\ 
$b_1$ & -3.31 & -3.59 & -2.25 \\ 
$b_2$ & -3.00 & -2.58 & -3.11 \\ 
$b_3$ & -3.89 & -3.48 & -3.62 \\ 
$b_4$ & -2.93 & -2.84 & -3.03 \\ 
$b_5$ & -4.58 & -3.50 & -3.26 \\ 
$b_6$ & -4.48 & -3.73 & -3.53 \\ 
$b_7$ & -2.86 & -3.27 & -1.87 \\ 
$b_8$ & -3.58 & -2.26 & -1.81 \\
\end{tabular}
\end{ruledtabular}
\end{table}

\section{Residuals}
\label{sec:app-errors}

In this section, plots are made of the residuals for each example using the FJet and RK models.
The residuals are calculated as
\begin{subequations}
\begin{align}
\text{res}(u) & = h_1 - \Delta u \\
\text{res}(\dot{u}) & = h_2 - \Delta \dot{u} 
\end{align}
\label{eqn:residuals}
\end{subequations}
where as before, $(\Delta u, \Delta \dot{u})$ are the data, and $(h_1, h_2)$ are the predictions.
For FJet, $h_1$ and $h_2$ are the derived models (cf. Eq.~\ref{eqn:h_model}); for RK2, 
they are the Taylor-expanded versions of the usual RK2 (cf. Eqs.~\ref{eqn:h-RK2-HO},\ref{eqn:h-RK2-pendulum},\ref{eqn:h-RK2-Duffing}).
Note that here the residual is defined as ``prediction - data", since the sampled data is taken as the ground truth.
(This is a minor point, but residuals are usually defined as ``data - prediction".)
Also, as before, the red (blue) dots indicate positive (negative) values, with the size of the dot being proportional to its magnitude.
Note that such plots are only conceivable for the FJet approach, and not the other approaches mentioned in Sec.~\ref{sec:background}.

Regions of predominantly blue or red dots indicate systematic deviations of the model from the data.
It may be due to a poor fit, or an insufficiency of terms in the feature set $X$ (i.e., a bias due to insufficient complexity in $X$).
Also, the maximum magnitudes of the residuals for each plot are summarized in Table~\ref{tab:residuals}; all values in that table are multiplied by $10^4$.

\begin{figure}
\begin{center}
\includegraphics[scale=0.25]{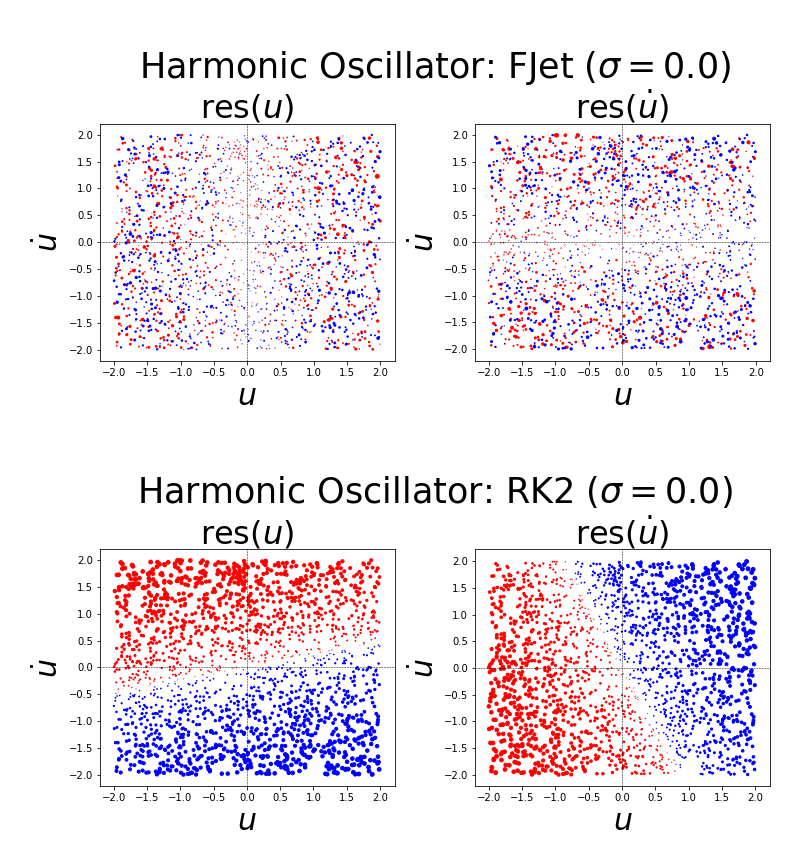}
\caption{Residuals for the harmonic oscillator example for the FJet model (top) and RK2 (bottom).  
The left and right plots are computed according to Eq.~\ref{eqn:residuals}.
The magnitude of the largest dots are given in Table~\ref{tab:residuals}.}
\label{fig:HO_residuals_g_n0}
\end{center}
\end{figure}

In each of the three examples, the figures were derived using the FJet model obtained when $\sigma=0$;
for the Duffing example, the $\sigma=0.1$ case was also included.
For the harmonic oscillator example shown in Fig.~\ref{fig:HO_residuals_g_n0}, the most noteworthy feature for the FJet model 
is the seemingly random placement of red and blue dots, which indicate no systematic residual.
As revealed in Table~\ref{tab:residuals}, the scale for the dots is extremely small, on the order of $10^{-15}$ (i.e., machine precision in this case).
These two facts are indicative of an excellent fit by FJet using feature regression.
In contrast, the RK2 residuals display systematic deviation, with dot sizes on a scale about $10^{11}$ times larger.

\begin{figure}
\begin{center}
\includegraphics[scale=0.25]{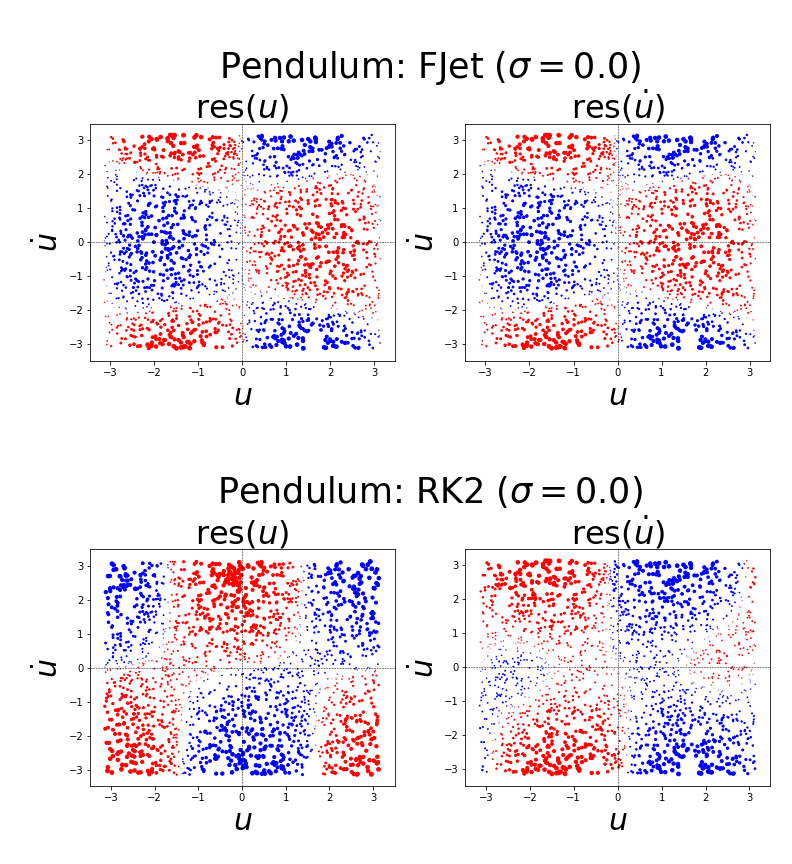}
\caption{Residuals for the pendulum example for the FJet model (top) and RK2 (bottom).  
The left and right plots are computed according to Eq.~\ref{eqn:residuals}.
The magnitude of the largest dots are given in Table~\ref{tab:residuals}.}
\label{fig:Pendulum_residuals_g_n0}
\end{center}
\end{figure}

The residuals for the pendulum example in Fig.~\ref{fig:Pendulum_residuals_g_n0} show systematic deviation for both FJet and RK2, and is likely
due to the limited feature set used in modeling.  
Higher order features are suggested in Appendix~\ref{sec:app-RK}.
As shown in Table~\ref{tab:residuals}, the residuals are smaller for FJet.

\begin{figure}
\begin{center}
\includegraphics[scale=0.25]{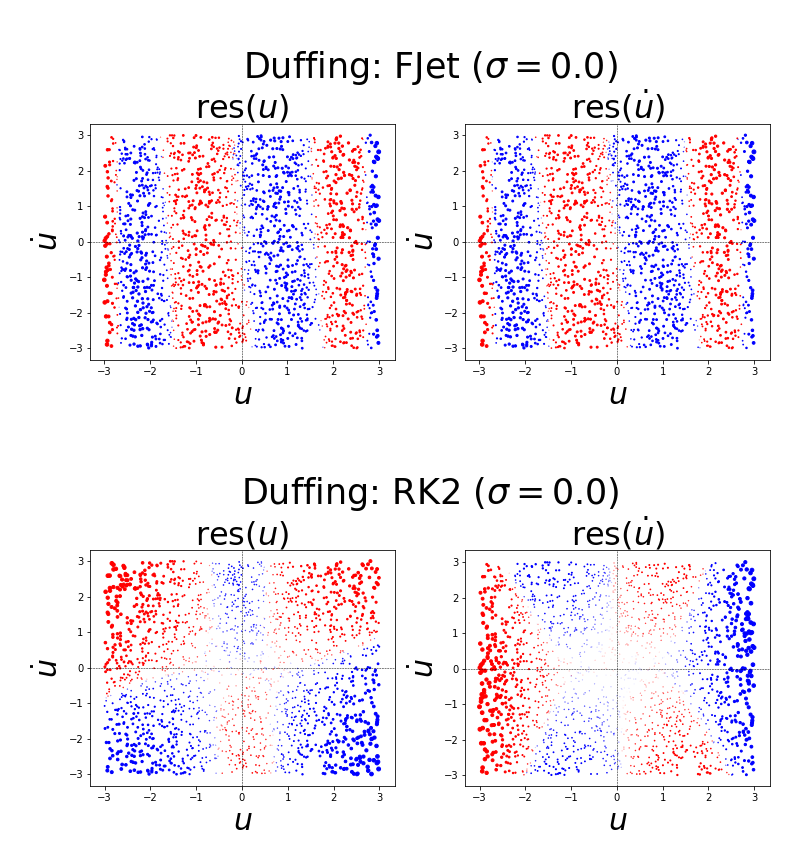}
\caption{Residuals for the Duffing oscillator example ($\sigma=0$) for the FJet model (top) and RK2 (bottom).  
The left and right plots are computed according to Eq.~\ref{eqn:residuals}.
The magnitude of the largest dots are given in Table~\ref{tab:residuals}.}
\label{fig:Duffing_residuals_g_n0}
\end{center}
\end{figure}

\begin{figure}
\begin{center}
\includegraphics[scale=0.25]{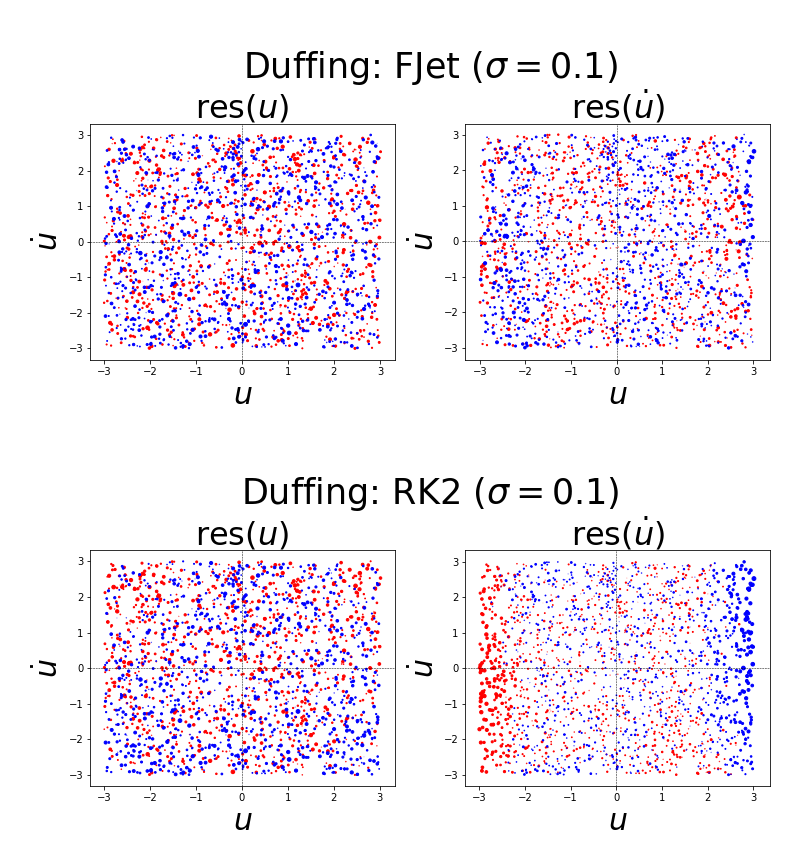}
\caption{Residuals for the Duffing oscillator example ($\sigma=0.1$) for the FJet model (top) and RK2 (bottom).  
The left and right plots are computed according to Eq.~\ref{eqn:residuals}.
The magnitude of the largest dots are given in Table~\ref{tab:residuals}.}
\label{fig:Duffing_residuals_g_n1}
\end{center}
\end{figure}

For the Duffing oscillator example, there are also systematic deviations seen in Fig.~\ref{fig:Duffing_residuals_g_n0}.
As with the Pendulum example, these are likely due to a limited feature set.
Figure \ref{fig:Duffing_residuals_g_n1} is also displayed, to show that when noise is present ($\sigma=0.1$) it can largely overwhelm the systematic deviations,
leaving mostly randomly placed red and blue dots.

\begin{table}[b]
\caption{\label{tab:residuals}%
This displays the {\em maximum} magnitudes of the residuals for the FJet and RK2 schemes, where $\text{mx}(u) = \text{max} | \text{res}(u) |$ and likewise for $\dot{u}$;
the maximum is taken over the data sample.
The first three examples use $\sigma=0$, while the last one uses $\sigma=0.1$.
All values were rounded and multiplied by $10^4$ to improve readability.
}
\begin{ruledtabular}
\begin{tabular}{lcccc}
 &\multicolumn{2}{c}{FJet } & \multicolumn{2}{c}{RK2}\\
  \textrm{Example}         &
  \textrm{$\text{mx}(u)$}         &
  \textrm{$\text{mx}(\dot{u})$}  &
  \textrm{$\text{mx}(u)$}      & 
  \textrm{$\text{mx}(\dot{u})$}    \\
\colrule
Harmonic Osc.     &  $3.9 \times 10^{-11}$ &  $3.1  \times 10^{-11}$  &  $3.9$  &  $4.4$ \\
Pendulum    &  $0.3$  &  $10.8$  &  $5.4$  &  $16.4$ \\
Duffing  &  $5$ &   $201$ &  $146$ &  $1069$ \\
Duffing ($0.1$)  & $471$ & $855$ &  $475$ & $1625$ \\
\end{tabular}
\end{ruledtabular}
\end{table}

\section{Second Optimization}
\label{sec:app-opt}

Taking a step back, the building of the model $h$ in Step \#3 of Sec.~\ref{sec:methodology} involved an optimization: minimizing the errors of the data $\Delta u$, $\Delta \dot{u}$ versus the model, taken over a sampling of $u$, $\dot{u}$ values.
However, having done that, it is possible to perform a second optimization, one in which the parameters are adjusted in order to minimize the error between a single, generated orbit and a corresponding orbit from data.
The cost function for this 
\begin{equation}
\sum_T \left\{
[ u_d(t)- \hat{u}(t) ]^2 + \alpha [ \dot{u}_d(t)- \hat{ \dot{u} }(t) ]^2
\right\}
\end{equation}
where ($u_d$, $\dot{u}_d$) represents the data, ($\hat{u}$, $\hat{\dot{u}}$) represents the solution generated from the model $h$, and $T$ signifies the set of times where the data is defined.  
(Recall that it is not required that the data be over regular time intervals.)
The reason this can lead to improvement is that the initial model fitting was premised on finding the best fit for a large set of possible values of $u$, $\dot{u}$.
This second optimization is instead focused more specifically on the errors near a particular orbit.
Note that the choice of $\alpha=1$ is appropriate to minimize the error in the energy, while $\alpha=0$ minimizes the error in $u$.
On a more practical note, the above cost function will be used with greedy updates formed from gaussian perturbations of the parameter values.
Finally, if this approach were to be generalized to an $n$th order ODE, one should include similar contributions up to the $(n-1)$st derivative.




\bibliography{../../../../Biblio/BIB_proj4_v3,../../../../Biblio/BIB_MZ,../../../../Biblio/BIB_ML,../../../../Biblio/BIB_math_Lie,../../../../Biblio/BIB_ML_Physics,../../../../Biblio/BIB_computer}

\begin{thebibliography}{70}%
\makeatletter
\providecommand \@ifxundefined [1]{%
 \@ifx{#1\undefined}
}%
\providecommand \@ifnum [1]{%
 \ifnum #1\expandafter \@firstoftwo
 \else \expandafter \@secondoftwo
 \fi
}%
\providecommand \@ifx [1]{%
 \ifx #1\expandafter \@firstoftwo
 \else \expandafter \@secondoftwo
 \fi
}%
\providecommand \natexlab [1]{#1}%
\providecommand \enquote  [1]{``#1''}%
\providecommand \bibnamefont  [1]{#1}%
\providecommand \bibfnamefont [1]{#1}%
\providecommand \citenamefont [1]{#1}%
\providecommand \href@noop [0]{\@secondoftwo}%
\providecommand \href [0]{\begingroup \@sanitize@url \@href}%
\providecommand \@href[1]{\@@startlink{#1}\@@href}%
\providecommand \@@href[1]{\endgroup#1\@@endlink}%
\providecommand \@sanitize@url [0]{\catcode `\\12\catcode `\$12\catcode
  `\&12\catcode `\#12\catcode `\^12\catcode `\_12\catcode `\%12\relax}%
\providecommand \@@startlink[1]{}%
\providecommand \@@endlink[0]{}%
\providecommand \url  [0]{\begingroup\@sanitize@url \@url }%
\providecommand \@url [1]{\endgroup\@href {#1}{\urlprefix }}%
\providecommand \urlprefix  [0]{URL }%
\providecommand \Eprint [0]{\href }%
\providecommand \doibase [0]{http://dx.doi.org/}%
\providecommand \selectlanguage [0]{\@gobble}%
\providecommand \bibinfo  [0]{\@secondoftwo}%
\providecommand \bibfield  [0]{\@secondoftwo}%
\providecommand \translation [1]{[#1]}%
\providecommand \BibitemOpen [0]{}%
\providecommand \bibitemStop [0]{}%
\providecommand \bibitemNoStop [0]{.\EOS\space}%
\providecommand \EOS [0]{\spacefactor3000\relax}%
\providecommand \BibitemShut  [1]{\csname bibitem#1\endcsname}%
\let\auto@bib@innerbib\@empty
\bibitem [{\citenamefont {Zimmer}(2021)}]{Zimmer-MLDE-arxiv}%
  \BibitemOpen
  \bibfield  {author} {\bibinfo {author} {\bibfnamefont {M.~F.}\ \bibnamefont
  {Zimmer}},\ }\href@noop {} {\bibfield  {journal} {\bibinfo  {journal} {arXiv
  preprint}\ }\textbf {\bibinfo {volume} {arXiv:2110.06917}} (\bibinfo {year}
  {2021})},\ \bibinfo {note} {(See Appendix C in version 2)}\BibitemShut
  {NoStop}%
\bibitem [{\citenamefont {Goldstein}(1980)}]{Goldstein-book-2nd}%
  \BibitemOpen
  \bibfield  {author} {\bibinfo {author} {\bibfnamefont {H.}~\bibnamefont
  {Goldstein}},\ }\href@noop {} {\emph {\bibinfo {title} {Classical
  Mechanics}}},\ \bibinfo {edition} {2nd}\ ed.\ (\bibinfo  {publisher}
  {Addison-Wesley},\ \bibinfo {year} {1980})\BibitemShut {NoStop}%
\bibitem [{\citenamefont {Galley}(2013)}]{Galley-2013}%
  \BibitemOpen
  \bibfield  {author} {\bibinfo {author} {\bibfnamefont {C.~R.}\ \bibnamefont
  {Galley}},\ }\href@noop {} {\bibfield  {journal} {\bibinfo  {journal}
  {Physical Review Letters}\ }\textbf {\bibinfo {volume} {110}} (\bibinfo
  {year} {2013})}\BibitemShut {NoStop}%
\bibitem [{\citenamefont {Poincar\'e}(1993)}]{Poincare-Goff-v1-1993}%
  \BibitemOpen
  \bibfield  {author} {\bibinfo {author} {\bibfnamefont {H.}~\bibnamefont
  {Poincar\'e}},\ }\href@noop {} {\emph {\bibinfo {title} {New Methods of
  Celestial Mechanics, 1: Periodic and Asymptotic Solutions \normalfont{(Ed.
  D.L. Goff)}}}}\ (\bibinfo  {publisher} {American Institute of Physics},\
  \bibinfo {year} {1993})\BibitemShut {NoStop}%
\bibitem [{\citenamefont {Holmes}(1990)}]{Holmes-1990}%
  \BibitemOpen
  \bibfield  {author} {\bibinfo {author} {\bibfnamefont {P.}~\bibnamefont
  {Holmes}},\ }\href@noop {} {\bibfield  {journal} {\bibinfo  {journal}
  {Physics Reports}\ }\textbf {\bibinfo {volume} {193}},\ \bibinfo {pages}
  {137} (\bibinfo {year} {1990})}\BibitemShut {NoStop}%
\bibitem [{\citenamefont {Poincar\'e}(1881)}]{Poincare-1881-DEs}%
  \BibitemOpen
  \bibfield  {author} {\bibinfo {author} {\bibfnamefont {H.}~\bibnamefont
  {Poincar\'e}},\ }\href@noop {} {\bibfield  {journal} {\bibinfo  {journal} {J.
  de Math.}\ }\textbf {\bibinfo {volume} {7}},\ \bibinfo {pages} {375}
  (\bibinfo {year} {1881})}\BibitemShut {NoStop}%
\bibitem [{\citenamefont {Lyapunov}(1992)}]{Lyapunov-book-reprint}%
  \BibitemOpen
  \bibfield  {author} {\bibinfo {author} {\bibfnamefont {A.~M.}\ \bibnamefont
  {Lyapunov}},\ }\href@noop {} {\emph {\bibinfo {title} {General Problem on
  Stability of Motion (English translation)}}}\ (\bibinfo  {publisher} {Taylor
  and Francis},\ \bibinfo {year} {1992})\ \bibinfo {note} {(Original work
  published 1892)}\BibitemShut {NoStop}%
\bibitem [{\citenamefont {Jackson}(1991{\natexlab{a}})}]{Atlee-book-v1}%
  \BibitemOpen
  \bibfield  {author} {\bibinfo {author} {\bibfnamefont {E.~A.}\ \bibnamefont
  {Jackson}},\ }\href@noop {} {\emph {\bibinfo {title} {Perspectives of
  nonlinear dynamics}}},\ Vol.~\bibinfo {volume} {1}\ (\bibinfo  {publisher}
  {Cambridge University Press},\ \bibinfo {year} {1991})\BibitemShut {NoStop}%
\bibitem [{\citenamefont {Sprott}(2003)}]{Sprott-book}%
  \BibitemOpen
  \bibfield  {author} {\bibinfo {author} {\bibfnamefont {J.~C.}\ \bibnamefont
  {Sprott}},\ }\href@noop {} {\emph {\bibinfo {title} {Chaos and Time-Series
  Analysis}}}\ (\bibinfo  {publisher} {Oxford},\ \bibinfo {year}
  {2003})\BibitemShut {NoStop}%
\bibitem [{\citenamefont {Lorenz}(1963)}]{Lorenz-1963}%
  \BibitemOpen
  \bibfield  {author} {\bibinfo {author} {\bibfnamefont {E.~N.}\ \bibnamefont
  {Lorenz}},\ }\href@noop {} {\bibfield  {journal} {\bibinfo  {journal}
  {Journal of Atmospheric Sciences}\ }\textbf {\bibinfo {volume} {20}},\
  \bibinfo {pages} {130} (\bibinfo {year} {1963})}\BibitemShut {NoStop}%
\bibitem [{\citenamefont {Devaney}(1986)}]{Devaney-book}%
  \BibitemOpen
  \bibfield  {author} {\bibinfo {author} {\bibfnamefont {R.~L.}\ \bibnamefont
  {Devaney}},\ }\href@noop {} {\emph {\bibinfo {title} {An Introduction to
  Chaotic Dynamical Systems}}}\ (\bibinfo  {publisher} {Benjamin/Cummings},\
  \bibinfo {year} {1986})\BibitemShut {NoStop}%
\bibitem [{\citenamefont {Jackson}(1991{\natexlab{b}})}]{Atlee-book-v2}%
  \BibitemOpen
  \bibfield  {author} {\bibinfo {author} {\bibfnamefont {E.~A.}\ \bibnamefont
  {Jackson}},\ }\href@noop {} {\emph {\bibinfo {title} {Perspectives of
  nonlinear dynamics}}},\ Vol.~\bibinfo {volume} {2}\ (\bibinfo  {publisher}
  {Cambridge University Press},\ \bibinfo {year} {1991})\BibitemShut {NoStop}%
\bibitem [{\citenamefont {Geist}\ \emph {et~al.}(1990)\citenamefont {Geist},
  \citenamefont {Parlitz},\ and\ \citenamefont {Born}}]{Geist-1990}%
  \BibitemOpen
  \bibfield  {author} {\bibinfo {author} {\bibfnamefont {K.}~\bibnamefont
  {Geist}}, \bibinfo {author} {\bibfnamefont {U.}~\bibnamefont {Parlitz}}, \
  and\ \bibinfo {author} {\bibfnamefont {W.~L.}\ \bibnamefont {Born}},\
  }\href@noop {} {\bibfield  {journal} {\bibinfo  {journal} {Progress of
  Theoretical Physics}\ }\textbf {\bibinfo {volume} {83}} (\bibinfo {year}
  {1990})}\BibitemShut {NoStop}%
\bibitem [{\citenamefont {Crutchfield}\ and\ \citenamefont
  {McNamara}(1987)}]{Crutchfield-1987}%
  \BibitemOpen
  \bibfield  {author} {\bibinfo {author} {\bibfnamefont {J.~P.}\ \bibnamefont
  {Crutchfield}}\ and\ \bibinfo {author} {\bibfnamefont {B.~S.}\ \bibnamefont
  {McNamara}},\ }\href@noop {} {\bibfield  {journal} {\bibinfo  {journal}
  {Complex Systems}\ }\textbf {\bibinfo {volume} {1}},\ \bibinfo {pages} {417}
  (\bibinfo {year} {1987})}\BibitemShut {NoStop}%
\bibitem [{\citenamefont {Bowen}(1978)}]{Bowen-1978}%
  \BibitemOpen
  \bibfield  {author} {\bibinfo {author} {\bibfnamefont {R.}~\bibnamefont
  {Bowen}},\ }\href@noop {} {\bibfield  {journal} {\bibinfo  {journal} {CBMS
  Regional Conference Series in Math.}\ }\textbf {\bibinfo {volume} {35}}
  (\bibinfo {year} {1978})}\BibitemShut {NoStop}%
\bibitem [{\citenamefont {Guckenheimer}\ and\ \citenamefont
  {Holmes}(1983)}]{Guckenheimer-book}%
  \BibitemOpen
  \bibfield  {author} {\bibinfo {author} {\bibfnamefont {J.}~\bibnamefont
  {Guckenheimer}}\ and\ \bibinfo {author} {\bibfnamefont {P.}~\bibnamefont
  {Holmes}},\ }\href@noop {} {\emph {\bibinfo {title} {Nonlinear Oscillations,
  Dynamical Systems, and Bifurcations of Vector Fields}}}\ (\bibinfo
  {publisher} {Springer-Verlag},\ \bibinfo {year} {1983})\BibitemShut {NoStop}%
\bibitem [{\citenamefont {McCauley}\ and\ \citenamefont
  {Palmore}(1986)}]{Palmore-1986}%
  \BibitemOpen
  \bibfield  {author} {\bibinfo {author} {\bibfnamefont {J.~L.}\ \bibnamefont
  {McCauley}}\ and\ \bibinfo {author} {\bibfnamefont {J.~I.}\ \bibnamefont
  {Palmore}},\ }\href@noop {} {\bibfield  {journal} {\bibinfo  {journal}
  {Physics Letters A}\ }\textbf {\bibinfo {volume} {15}},\ \bibinfo {pages}
  {433} (\bibinfo {year} {1986})}\BibitemShut {NoStop}%
\bibitem [{\citenamefont {Palmore}\ and\ \citenamefont
  {McCauley}(1987)}]{Palmore-1987}%
  \BibitemOpen
  \bibfield  {author} {\bibinfo {author} {\bibfnamefont {J.~I.}\ \bibnamefont
  {Palmore}}\ and\ \bibinfo {author} {\bibfnamefont {J.~L.}\ \bibnamefont
  {McCauley}},\ }\href@noop {} {\bibfield  {journal} {\bibinfo  {journal}
  {Physics Letters A}\ }\textbf {\bibinfo {volume} {122}},\ \bibinfo {pages}
  {399} (\bibinfo {year} {1987})}\BibitemShut {NoStop}%
\bibitem [{\citenamefont {Grebogi}\ \emph {et~al.}(1990)\citenamefont
  {Grebogi}, \citenamefont {Hammel}, \citenamefont {Yorke},\ and\ \citenamefont
  {Sauer}}]{Grebogi-1990}%
  \BibitemOpen
  \bibfield  {author} {\bibinfo {author} {\bibfnamefont {C.}~\bibnamefont
  {Grebogi}}, \bibinfo {author} {\bibfnamefont {S.~M.}\ \bibnamefont {Hammel}},
  \bibinfo {author} {\bibfnamefont {J.~A.}\ \bibnamefont {Yorke}}, \ and\
  \bibinfo {author} {\bibfnamefont {T.}~\bibnamefont {Sauer}},\ }\href@noop {}
  {\bibfield  {journal} {\bibinfo  {journal} {Phys. Rev. Lett.}\ }\textbf
  {\bibinfo {volume} {65}} (\bibinfo {year} {1990})}\BibitemShut {NoStop}%
\bibitem [{\citenamefont {Sauer}\ \emph {et~al.}(1997)\citenamefont {Sauer},
  \citenamefont {Grebogi},\ and\ \citenamefont {Yorke}}]{Sauer-1997}%
  \BibitemOpen
  \bibfield  {author} {\bibinfo {author} {\bibfnamefont {T.}~\bibnamefont
  {Sauer}}, \bibinfo {author} {\bibfnamefont {C.}~\bibnamefont {Grebogi}}, \
  and\ \bibinfo {author} {\bibfnamefont {J.~A.}\ \bibnamefont {Yorke}},\
  }\href@noop {} {\bibfield  {journal} {\bibinfo  {journal} {Phys. Rev. Lett.}\
  }\textbf {\bibinfo {volume} {79}} (\bibinfo {year} {1997})}\BibitemShut
  {NoStop}%
\bibitem [{\citenamefont {Ruelle}\ and\ \citenamefont
  {Takens}(1971)}]{Ruelle-1971}%
  \BibitemOpen
  \bibfield  {author} {\bibinfo {author} {\bibfnamefont {D.}~\bibnamefont
  {Ruelle}}\ and\ \bibinfo {author} {\bibfnamefont {F.}~\bibnamefont
  {Takens}},\ }\href@noop {} {\bibfield  {journal} {\bibinfo  {journal} {Comm.
  Math. Phys.}\ }\textbf {\bibinfo {volume} {20}} (\bibinfo {year}
  {1971})}\BibitemShut {NoStop}%
\bibitem [{\citenamefont {Packard}\ \emph {et~al.}(1980)\citenamefont
  {Packard}, \citenamefont {Crutchfield}, \citenamefont {Farmer},\ and\
  \citenamefont {Shaw}}]{Packard-1980-geometry}%
  \BibitemOpen
  \bibfield  {author} {\bibinfo {author} {\bibfnamefont {N.~H.}\ \bibnamefont
  {Packard}}, \bibinfo {author} {\bibfnamefont {J.~P.}\ \bibnamefont
  {Crutchfield}}, \bibinfo {author} {\bibfnamefont {J.~D.}\ \bibnamefont
  {Farmer}}, \ and\ \bibinfo {author} {\bibfnamefont {R.~S.}\ \bibnamefont
  {Shaw}},\ }\href@noop {} {\bibfield  {journal} {\bibinfo  {journal} {Physical
  Review Letters}\ }\textbf {\bibinfo {volume} {45}} (\bibinfo {year}
  {1980})}\BibitemShut {NoStop}%
\bibitem [{\citenamefont {Takens}(1981)}]{Takens-1981}%
  \BibitemOpen
  \bibfield  {author} {\bibinfo {author} {\bibfnamefont {F.}~\bibnamefont
  {Takens}},\ }in\ \href@noop {} {\emph {\bibinfo {booktitle} {Lecture Notes in
  Mathematics: Dynamical Systems and Turbulence, Warwick 1980 (Coventry,
  1979/1980)}}},\ Vol.\ \bibinfo {volume} {898}\ (\bibinfo  {publisher}
  {Springer},\ \bibinfo {year} {1981})\ pp.\ \bibinfo {pages}
  {366--381}\BibitemShut {NoStop}%
\bibitem [{\citenamefont {Farmer}\ and\ \citenamefont
  {Sidorowich}(1987)}]{Farmer-1987}%
  \BibitemOpen
  \bibfield  {author} {\bibinfo {author} {\bibfnamefont {J.~D.}\ \bibnamefont
  {Farmer}}\ and\ \bibinfo {author} {\bibfnamefont {J.~J.}\ \bibnamefont
  {Sidorowich}},\ }\href@noop {} {\bibfield  {journal} {\bibinfo  {journal}
  {Physical Review Letters}\ }\textbf {\bibinfo {volume} {59}},\ \bibinfo
  {pages} {845} (\bibinfo {year} {1987})}\BibitemShut {NoStop}%
\bibitem [{\citenamefont {Roux}\ \emph {et~al.}(1980)\citenamefont {Roux},
  \citenamefont {Rossi}, \citenamefont {Bachelart},\ and\ \citenamefont
  {Vidal}}]{Roux-1980}%
  \BibitemOpen
  \bibfield  {author} {\bibinfo {author} {\bibfnamefont {J.~C.}\ \bibnamefont
  {Roux}}, \bibinfo {author} {\bibfnamefont {A.}~\bibnamefont {Rossi}},
  \bibinfo {author} {\bibfnamefont {S.}~\bibnamefont {Bachelart}}, \ and\
  \bibinfo {author} {\bibfnamefont {C.}~\bibnamefont {Vidal}},\ }\href@noop {}
  {\bibfield  {journal} {\bibinfo  {journal} {Phys. Lett. A}\ }\textbf
  {\bibinfo {volume} {77}},\ \bibinfo {pages} {391} (\bibinfo {year}
  {1980})}\BibitemShut {NoStop}%
\bibitem [{\citenamefont {Brandstater}\ \emph {et~al.}(1980)\citenamefont
  {Brandstater}, \citenamefont {Swift}, \citenamefont {Swinney}, \citenamefont
  {Wolf}, \citenamefont {Farmer}, \citenamefont {Jen},\ and\ \citenamefont
  {Crutchfield}}]{Brandstater-1983}%
  \BibitemOpen
  \bibfield  {author} {\bibinfo {author} {\bibfnamefont {A.}~\bibnamefont
  {Brandstater}}, \bibinfo {author} {\bibfnamefont {J.}~\bibnamefont {Swift}},
  \bibinfo {author} {\bibfnamefont {H.~L.}\ \bibnamefont {Swinney}}, \bibinfo
  {author} {\bibfnamefont {A.}~\bibnamefont {Wolf}}, \bibinfo {author}
  {\bibfnamefont {J.~D.}\ \bibnamefont {Farmer}}, \bibinfo {author}
  {\bibfnamefont {E.}~\bibnamefont {Jen}}, \ and\ \bibinfo {author}
  {\bibfnamefont {J.~P.}\ \bibnamefont {Crutchfield}},\ }\href@noop {}
  {\bibfield  {journal} {\bibinfo  {journal} {Phys. Lett. A}\ }\textbf
  {\bibinfo {volume} {77}},\ \bibinfo {pages} {391} (\bibinfo {year}
  {1980})}\BibitemShut {NoStop}%
\bibitem [{\citenamefont {Broomhead}\ and\ \citenamefont
  {Lowe}(1988)}]{Broomhead-1988}%
  \BibitemOpen
  \bibfield  {author} {\bibinfo {author} {\bibfnamefont {D.~S.}\ \bibnamefont
  {Broomhead}}\ and\ \bibinfo {author} {\bibfnamefont {D.}~\bibnamefont
  {Lowe}},\ }\href@noop {} {\bibfield  {journal} {\bibinfo  {journal} {Complex
  Systems}\ }\textbf {\bibinfo {volume} {2}},\ \bibinfo {pages} {321} (\bibinfo
  {year} {1988})}\BibitemShut {NoStop}%
\bibitem [{\citenamefont {Casdagli}(1989)}]{Casdagli-1989}%
  \BibitemOpen
  \bibfield  {author} {\bibinfo {author} {\bibfnamefont {M.}~\bibnamefont
  {Casdagli}},\ }\href@noop {} {\bibfield  {journal} {\bibinfo  {journal}
  {Physica D}\ }\textbf {\bibinfo {volume} {35}},\ \bibinfo {pages} {335}
  (\bibinfo {year} {1989})}\BibitemShut {NoStop}%
\bibitem [{\citenamefont {Cremers}\ and\ \citenamefont
  {H\"ubler}(1987)}]{Cremers-Hubler-1987}%
  \BibitemOpen
  \bibfield  {author} {\bibinfo {author} {\bibfnamefont {J.}~\bibnamefont
  {Cremers}}\ and\ \bibinfo {author} {\bibfnamefont {A.}~\bibnamefont
  {H\"ubler}},\ }\href@noop {} {\bibfield  {journal} {\bibinfo  {journal}
  {Zeitschrift f\"ur Naturforschung A}\ }\textbf {\bibinfo {volume} {42}},\
  \bibinfo {pages} {797} (\bibinfo {year} {1987})}\BibitemShut {NoStop}%
\bibitem [{\citenamefont {Breeden}\ and\ \citenamefont
  {Hübler}(1990)}]{Breeden-1990}%
  \BibitemOpen
  \bibfield  {author} {\bibinfo {author} {\bibfnamefont {J.~L.}\ \bibnamefont
  {Breeden}}\ and\ \bibinfo {author} {\bibfnamefont {A.}~\bibnamefont
  {Hübler}},\ }\href@noop {} {\bibfield  {journal} {\bibinfo  {journal}
  {Physical Review A}\ }\textbf {\bibinfo {volume} {42}} (\bibinfo {year}
  {1990})}\BibitemShut {NoStop}%
\bibitem [{\citenamefont {Gouesbet}(1991)}]{Gouesbet-1991}%
  \BibitemOpen
  \bibfield  {author} {\bibinfo {author} {\bibfnamefont {G.}~\bibnamefont
  {Gouesbet}},\ }\href@noop {} {\bibfield  {journal} {\bibinfo  {journal}
  {Physical Review A}\ }\textbf {\bibinfo {volume} {43}},\ \bibinfo {pages}
  {5321} (\bibinfo {year} {1991})}\BibitemShut {NoStop}%
\bibitem [{\citenamefont {Gouesbet}(1992)}]{Gouesbet-1992}%
  \BibitemOpen
  \bibfield  {author} {\bibinfo {author} {\bibfnamefont {G.}~\bibnamefont
  {Gouesbet}},\ }\href@noop {} {\bibfield  {journal} {\bibinfo  {journal}
  {Physical Review A}\ }\textbf {\bibinfo {volume} {46}},\ \bibinfo {pages}
  {1784–1796} (\bibinfo {year} {1992})}\BibitemShut {NoStop}%
\bibitem [{\citenamefont {Broomhead}\ \emph {et~al.}(1991)\citenamefont
  {Broomhead}, \citenamefont {Indik}, \citenamefont {Newell},\ and\
  \citenamefont {Rand}}]{Broomhead-1991}%
  \BibitemOpen
  \bibfield  {author} {\bibinfo {author} {\bibfnamefont {D.~S.}\ \bibnamefont
  {Broomhead}}, \bibinfo {author} {\bibfnamefont {R.}~\bibnamefont {Indik}},
  \bibinfo {author} {\bibfnamefont {A.~C.}\ \bibnamefont {Newell}}, \ and\
  \bibinfo {author} {\bibfnamefont {D.~A.}\ \bibnamefont {Rand}},\ }\href@noop
  {} {\bibfield  {journal} {\bibinfo  {journal} {Nonlinearity}\ }\textbf
  {\bibinfo {volume} {4}} (\bibinfo {year} {1991})}\BibitemShut {NoStop}%
\bibitem [{\citenamefont {Aguirre}\ and\ \citenamefont
  {Letellier}(2009)}]{Aguirre-review-2009}%
  \BibitemOpen
  \bibfield  {author} {\bibinfo {author} {\bibfnamefont {L.~A.}\ \bibnamefont
  {Aguirre}}\ and\ \bibinfo {author} {\bibfnamefont {C.}~\bibnamefont
  {Letellier}},\ }\href@noop {} {\bibfield  {journal} {\bibinfo  {journal}
  {Mathematical Problems in Engineering}\ }\textbf {\bibinfo {volume} {2009}}
  (\bibinfo {year} {2009})}\BibitemShut {NoStop}%
\bibitem [{\citenamefont {Greydanus}\ \emph {et~al.}(2019)\citenamefont
  {Greydanus}, \citenamefont {Dzamba},\ and\ \citenamefont
  {Yosinski}}]{Greydanus-HNN-2019}%
  \BibitemOpen
  \bibfield  {author} {\bibinfo {author} {\bibfnamefont {S.}~\bibnamefont
  {Greydanus}}, \bibinfo {author} {\bibfnamefont {M.}~\bibnamefont {Dzamba}}, \
  and\ \bibinfo {author} {\bibfnamefont {J.}~\bibnamefont {Yosinski}},\ }in\
  \href@noop {} {\emph {\bibinfo {booktitle} {Neural Information Processing
  Systems (NeurIPS)}}}\ (\bibinfo {year} {2019})\ pp.\ \bibinfo {pages}
  {15353--15363}\BibitemShut {NoStop}%
\bibitem [{\citenamefont {Bondesan}\ and\ \citenamefont
  {Lamacraft}(2019)}]{Bondesan-2019}%
  \BibitemOpen
  \bibfield  {author} {\bibinfo {author} {\bibfnamefont {R.}~\bibnamefont
  {Bondesan}}\ and\ \bibinfo {author} {\bibfnamefont {A.}~\bibnamefont
  {Lamacraft}},\ }\href@noop {} {\bibfield  {journal} {\bibinfo  {journal}
  {arXiv preprint}\ }\textbf {\bibinfo {volume} {arXiv:1906.04645}} (\bibinfo
  {year} {2019})},\ \bibinfo {note} {(Presented at ICML 2019 Workshop on
  Theoretical Physics for Deep Learning.)}\BibitemShut {NoStop}%
\bibitem [{\citenamefont {Toth}\ \emph {et~al.}(2019)\citenamefont {Toth},
  \citenamefont {Rezende}, \citenamefont {Jaegle}, \citenamefont {Racanière},
  \citenamefont {Botev},\ and\ \citenamefont {Higgins}}]{Toth-2020}%
  \BibitemOpen
  \bibfield  {author} {\bibinfo {author} {\bibfnamefont {P.}~\bibnamefont
  {Toth}}, \bibinfo {author} {\bibfnamefont {D.~J.}\ \bibnamefont {Rezende}},
  \bibinfo {author} {\bibfnamefont {A.}~\bibnamefont {Jaegle}}, \bibinfo
  {author} {\bibfnamefont {S.}~\bibnamefont {Racanière}}, \bibinfo {author}
  {\bibfnamefont {A.}~\bibnamefont {Botev}}, \ and\ \bibinfo {author}
  {\bibfnamefont {I.}~\bibnamefont {Higgins}}\ }(\bibinfo {year}
  {2019})\BibitemShut {NoStop}%
\bibitem [{\citenamefont {Lutter}\ \emph {et~al.}(2019)\citenamefont {Lutter},
  \citenamefont {Ritter},\ and\ \citenamefont {Peters}}]{Lutter-DeLaN-2019}%
  \BibitemOpen
  \bibfield  {author} {\bibinfo {author} {\bibfnamefont {M.}~\bibnamefont
  {Lutter}}, \bibinfo {author} {\bibfnamefont {C.}~\bibnamefont {Ritter}}, \
  and\ \bibinfo {author} {\bibfnamefont {J.}~\bibnamefont {Peters}},\ }in\
  \href@noop {} {\emph {\bibinfo {booktitle} {International Conference on
  Learning Representations (ICLR)}}}\ (\bibinfo {year} {2019})\BibitemShut
  {NoStop}%
\bibitem [{\citenamefont {Cranmer}\ \emph
  {et~al.}(2020{\natexlab{a}})\citenamefont {Cranmer}, \citenamefont
  {Greydanus}, \citenamefont {Hoyer}, \citenamefont {Battaglia}, \citenamefont
  {Spergel},\ and\ \citenamefont {Ho}}]{Cranmer-LNN-2020}%
  \BibitemOpen
  \bibfield  {author} {\bibinfo {author} {\bibfnamefont {M.}~\bibnamefont
  {Cranmer}}, \bibinfo {author} {\bibfnamefont {S.}~\bibnamefont {Greydanus}},
  \bibinfo {author} {\bibfnamefont {S.}~\bibnamefont {Hoyer}}, \bibinfo
  {author} {\bibfnamefont {P.}~\bibnamefont {Battaglia}}, \bibinfo {author}
  {\bibfnamefont {D.}~\bibnamefont {Spergel}}, \ and\ \bibinfo {author}
  {\bibfnamefont {S.}~\bibnamefont {Ho}},\ }in\ \href@noop {} {\emph {\bibinfo
  {booktitle} {ICLR 2020 Workshop on Integration of Deep Neural Models and
  Differential Equations}}}\ (\bibinfo {year} {2020})\BibitemShut {NoStop}%
\bibitem [{\citenamefont {Cranmer}\ \emph
  {et~al.}(2020{\natexlab{b}})\citenamefont {Cranmer}, \citenamefont
  {Sanchez-Gonzalez}, \citenamefont {Battaglia}, \citenamefont {Xu},
  \citenamefont {Cranmer}, \citenamefont {Spergel},\ and\ \citenamefont
  {Ho}}]{Cranmer-2020}%
  \BibitemOpen
  \bibfield  {author} {\bibinfo {author} {\bibfnamefont {M.}~\bibnamefont
  {Cranmer}}, \bibinfo {author} {\bibfnamefont {A.}~\bibnamefont
  {Sanchez-Gonzalez}}, \bibinfo {author} {\bibfnamefont {P.}~\bibnamefont
  {Battaglia}}, \bibinfo {author} {\bibfnamefont {R.}~\bibnamefont {Xu}},
  \bibinfo {author} {\bibfnamefont {K.}~\bibnamefont {Cranmer}}, \bibinfo
  {author} {\bibfnamefont {D.}~\bibnamefont {Spergel}}, \ and\ \bibinfo
  {author} {\bibfnamefont {S.}~\bibnamefont {Ho}},\ }in\ \href@noop {} {\emph
  {\bibinfo {booktitle} {Advances in Neural Information Processing Systems 33
  (NeurIPS)}}}\ (\bibinfo {year} {2020})\BibitemShut {NoStop}%
\bibitem [{\citenamefont {Rudy}\ \emph {et~al.}(2017)\citenamefont {Rudy},
  \citenamefont {Brunton}, \citenamefont {Proctor},\ and\ \citenamefont
  {Kutz}}]{Rudy-2017}%
  \BibitemOpen
  \bibfield  {author} {\bibinfo {author} {\bibfnamefont {S.~H.}\ \bibnamefont
  {Rudy}}, \bibinfo {author} {\bibfnamefont {S.~L.}\ \bibnamefont {Brunton}},
  \bibinfo {author} {\bibfnamefont {J.~L.}\ \bibnamefont {Proctor}}, \ and\
  \bibinfo {author} {\bibfnamefont {J.~N.}\ \bibnamefont {Kutz}},\ }\href@noop
  {} {\bibfield  {journal} {\bibinfo  {journal} {Science Advances}\ }\textbf
  {\bibinfo {volume} {3}} (\bibinfo {year} {2017})}\BibitemShut {NoStop}%
\bibitem [{\citenamefont {Chen}\ \emph {et~al.}(2018)\citenamefont {Chen},
  \citenamefont {Rubanova}, \citenamefont {Bettencourt},\ and\ \citenamefont
  {Duvenaud}}]{Chen-2018}%
  \BibitemOpen
  \bibfield  {author} {\bibinfo {author} {\bibfnamefont {R.~T.~Q.}\
  \bibnamefont {Chen}}, \bibinfo {author} {\bibfnamefont {Y.}~\bibnamefont
  {Rubanova}}, \bibinfo {author} {\bibfnamefont {J.}~\bibnamefont
  {Bettencourt}}, \ and\ \bibinfo {author} {\bibfnamefont {D.}~\bibnamefont
  {Duvenaud}},\ }in\ \href@noop {} {\emph {\bibinfo {booktitle} {Neural
  Information Processing Systems (NeurIPS)}}}\ (\bibinfo {year} {2018})\ pp.\
  \bibinfo {pages} {6572--6583}\BibitemShut {NoStop}%
\bibitem [{\citenamefont {Ott}\ \emph {et~al.}(2021)\citenamefont {Ott},
  \citenamefont {Katiyar}, \citenamefont {Hennig},\ and\ \citenamefont
  {Tiemann}}]{Ott-2020}%
  \BibitemOpen
  \bibfield  {author} {\bibinfo {author} {\bibfnamefont {K.}~\bibnamefont
  {Ott}}, \bibinfo {author} {\bibfnamefont {P.}~\bibnamefont {Katiyar}},
  \bibinfo {author} {\bibfnamefont {P.}~\bibnamefont {Hennig}}, \ and\ \bibinfo
  {author} {\bibfnamefont {M.}~\bibnamefont {Tiemann}},\ }in\ \href@noop {}
  {\emph {\bibinfo {booktitle} {International Conference on Learning
  Representations (ICLR)}}}\ (\bibinfo {year} {2021})\BibitemShut {NoStop}%
\bibitem [{\citenamefont {Kidger}(2022)}]{Kidger-2022}%
  \BibitemOpen
  \bibfield  {author} {\bibinfo {author} {\bibfnamefont {P.}~\bibnamefont
  {Kidger}},\ }\href@noop {} {\bibfield  {journal} {\bibinfo  {journal} {arXiv
  preprint}\ }\textbf {\bibinfo {volume} {arXiv:2202.02435}} (\bibinfo {year}
  {2022})}\BibitemShut {NoStop}%
\bibitem [{\citenamefont {Box}\ \emph {et~al.}(2015)\citenamefont {Box},
  \citenamefont {Jenkins}, \citenamefont {Reinsel},\ and\ \citenamefont
  {Ljung}}]{BoxJenkins-5th}%
  \BibitemOpen
  \bibfield  {author} {\bibinfo {author} {\bibfnamefont {G.~E.~P.}\
  \bibnamefont {Box}}, \bibinfo {author} {\bibfnamefont {G.~M.}\ \bibnamefont
  {Jenkins}}, \bibinfo {author} {\bibfnamefont {G.~C.}\ \bibnamefont
  {Reinsel}}, \ and\ \bibinfo {author} {\bibfnamefont {G.~M.}\ \bibnamefont
  {Ljung}},\ }\href@noop {} {\emph {\bibinfo {title} {Time Series Analysis:
  Forecasting and Control}}},\ \bibinfo {edition} {5th}\ ed.\ (\bibinfo
  {publisher} {Wiley},\ \bibinfo {year} {2015})\BibitemShut {NoStop}%
\bibitem [{\citenamefont {Hairer}\ \emph {et~al.}(1987)\citenamefont {Hairer},
  \citenamefont {Nørsett},\ and\ \citenamefont {Wanner}}]{Hairer-1987}%
  \BibitemOpen
  \bibfield  {author} {\bibinfo {author} {\bibfnamefont {E.}~\bibnamefont
  {Hairer}}, \bibinfo {author} {\bibfnamefont {S.~P.}\ \bibnamefont
  {Nørsett}}, \ and\ \bibinfo {author} {\bibfnamefont {G.}~\bibnamefont
  {Wanner}},\ }\href@noop {} {\emph {\bibinfo {title} {Solving Ordinary
  Differential Equations I – Nonstiff Problems}}}\ (\bibinfo  {publisher}
  {Springer},\ \bibinfo {year} {1987})\BibitemShut {NoStop}%
\bibitem [{\citenamefont {Press}\ \emph {et~al.}(2007)\citenamefont {Press},
  \citenamefont {Teukolsky}, \citenamefont {Vetterling},\ and\ \citenamefont
  {Flannery}}]{numrec}%
  \BibitemOpen
  \bibfield  {author} {\bibinfo {author} {\bibfnamefont {W.~H.}\ \bibnamefont
  {Press}}, \bibinfo {author} {\bibfnamefont {S.~A.}\ \bibnamefont
  {Teukolsky}}, \bibinfo {author} {\bibfnamefont {W.~T.}\ \bibnamefont
  {Vetterling}}, \ and\ \bibinfo {author} {\bibfnamefont {B.~P.}\ \bibnamefont
  {Flannery}},\ }\href@noop {} {\emph {\bibinfo {title} {Numerical Recipes, The
  Art of Scientific Computing}}},\ \bibinfo {edition} {3rd}\ ed.\ (\bibinfo
  {publisher} {Cambridge University Press},\ \bibinfo {year}
  {2007})\BibitemShut {NoStop}%
\bibitem [{\citenamefont {Schober}\ \emph {et~al.}(2014)\citenamefont
  {Schober}, \citenamefont {Duvenaud},\ and\ \citenamefont
  {Hennig}}]{Schober-2014}%
  \BibitemOpen
  \bibfield  {author} {\bibinfo {author} {\bibfnamefont {M.}~\bibnamefont
  {Schober}}, \bibinfo {author} {\bibfnamefont {D.}~\bibnamefont {Duvenaud}}, \
  and\ \bibinfo {author} {\bibfnamefont {P.}~\bibnamefont {Hennig}},\ }in\
  \href@noop {} {\emph {\bibinfo {booktitle} {Neural Information Processing
  Systems (NeurIPS)}}}\ (\bibinfo {year} {2014})\BibitemShut {NoStop}%
\bibitem [{\citenamefont {Ince}(1956)}]{Ince-1956}%
  \BibitemOpen
  \bibfield  {author} {\bibinfo {author} {\bibfnamefont {E.~L.}\ \bibnamefont
  {Ince}},\ }\href@noop {} {\emph {\bibinfo {title} {Ordinary Differential
  Equations}}}\ (\bibinfo  {publisher} {Dover},\ \bibinfo {year}
  {1956})\BibitemShut {NoStop}%
\bibitem [{\citenamefont {Olver}(1993)}]{Olver-1993}%
  \BibitemOpen
  \bibfield  {author} {\bibinfo {author} {\bibfnamefont {P.~J.}\ \bibnamefont
  {Olver}},\ }\href@noop {} {\emph {\bibinfo {title} {Applications of Lie
  Groups to Differential Equations}}},\ \bibinfo {edition} {2nd}\ ed.\
  (\bibinfo  {publisher} {Springer-Verlag},\ \bibinfo {year}
  {1993})\BibitemShut {NoStop}%
\bibitem [{\citenamefont {Liu}\ and\ \citenamefont
  {Tegmark}(2021)}]{ZLiu-2021}%
  \BibitemOpen
  \bibfield  {author} {\bibinfo {author} {\bibfnamefont {Z.}~\bibnamefont
  {Liu}}\ and\ \bibinfo {author} {\bibfnamefont {M.}~\bibnamefont {Tegmark}},\
  }\href@noop {} {\bibfield  {journal} {\bibinfo  {journal} {Phys. Rev. Lett.}\
  }\textbf {\bibinfo {volume} {126}},\ \bibinfo {pages} {180604} (\bibinfo
  {year} {2021})}\BibitemShut {NoStop}%
\bibitem [{\citenamefont {Mototake}(2021)}]{YMototake-2021}%
  \BibitemOpen
  \bibfield  {author} {\bibinfo {author} {\bibfnamefont {Y.}~\bibnamefont
  {Mototake}},\ }\href@noop {} {\bibfield  {journal} {\bibinfo  {journal}
  {Phys. Rev. E}\ }\textbf {\bibinfo {volume} {103}},\ \bibinfo {pages}
  {033303} (\bibinfo {year} {2021})}\BibitemShut {NoStop}%
\bibitem [{\citenamefont {Knowles}\ and\ \citenamefont
  {Renka}(2014)}]{Knowles-2014}%
  \BibitemOpen
  \bibfield  {author} {\bibinfo {author} {\bibfnamefont {I.}~\bibnamefont
  {Knowles}}\ and\ \bibinfo {author} {\bibfnamefont {R.~J.}\ \bibnamefont
  {Renka}},\ }\href@noop {} {\bibfield  {journal} {\bibinfo  {journal}
  {Electronic Journal of Differential Equations, Conference}\ }\textbf
  {\bibinfo {volume} {21}},\ \bibinfo {pages} {235–246} (\bibinfo {year}
  {2014})}\BibitemShut {NoStop}%
\bibitem [{\citenamefont {Letellier}\ \emph {et~al.}(2009)\citenamefont
  {Letellier}, \citenamefont {Aguirre},\ and\ \citenamefont
  {Freitas}}]{Letellier-2009}%
  \BibitemOpen
  \bibfield  {author} {\bibinfo {author} {\bibfnamefont {C.}~\bibnamefont
  {Letellier}}, \bibinfo {author} {\bibfnamefont {L.~A.}\ \bibnamefont
  {Aguirre}}, \ and\ \bibinfo {author} {\bibfnamefont {U.~S.}\ \bibnamefont
  {Freitas}},\ }\href@noop {} {\bibfield  {journal} {\bibinfo  {journal}
  {Chaos}\ }\textbf {\bibinfo {volume} {19}} (\bibinfo {year}
  {2009})}\BibitemShut {NoStop}%
\bibitem [{\citenamefont {Wiener}(1950)}]{Wiener-1950}%
  \BibitemOpen
  \bibfield  {author} {\bibinfo {author} {\bibfnamefont {N.}~\bibnamefont
  {Wiener}},\ }\href@noop {} {\bibfield  {journal} {\bibinfo  {journal} {Bull.
  Amer. Math. Soc.}\ }\textbf {\bibinfo {volume} {56}},\ \bibinfo {pages}
  {378–381} (\bibinfo {year} {1950})}\BibitemShut {NoStop}%
\bibitem [{\citenamefont {Hastie}\ \emph {et~al.}(2009)\citenamefont {Hastie},
  \citenamefont {Tibshirani},\ and\ \citenamefont {Friedman}}]{Hastie_ESL2}%
  \BibitemOpen
  \bibfield  {author} {\bibinfo {author} {\bibfnamefont {T.}~\bibnamefont
  {Hastie}}, \bibinfo {author} {\bibfnamefont {R.}~\bibnamefont {Tibshirani}},
  \ and\ \bibinfo {author} {\bibfnamefont {J.}~\bibnamefont {Friedman}},\
  }\href@noop {} {\emph {\bibinfo {title} {The Elements of Statistical
  Learning: Data Mining, Inference, and Prediction}}},\ \bibinfo {edition}
  {2nd}\ ed.\ (\bibinfo  {publisher} {Springer-Verlag},\ \bibinfo {year}
  {2009})\BibitemShut {NoStop}%
\bibitem [{\citenamefont {Young}(2020)}]{Young_notes}%
  \BibitemOpen
  \bibfield  {author} {\bibinfo {author} {\bibfnamefont {A.~P.}\ \bibnamefont
  {Young}},\ }\href {https://young.physics.ucsc.edu/115/ode_solve.pdf}
  {\enquote {\bibinfo {title} {Physics 115/242, {C}omparison of methods for
  integrating the simple harmonic oscillator},}\ } (\bibinfo {year}
  {2020})\BibitemShut {NoStop}%
\bibitem [{\citenamefont {Duffing}(1918)}]{Duffing-1918}%
  \BibitemOpen
  \bibfield  {author} {\bibinfo {author} {\bibfnamefont {G.}~\bibnamefont
  {Duffing}},\ }\href@noop {} {\bibfield  {journal} {\bibinfo  {journal}
  {Vieweg, Braunschweig}\ }\textbf {\bibinfo {volume} {41/42}} (\bibinfo {year}
  {1918})}\BibitemShut {NoStop}%
\bibitem [{\citenamefont {Holmes}\ and\ \citenamefont
  {Rand}(1976)}]{Holmes-1976}%
  \BibitemOpen
  \bibfield  {author} {\bibinfo {author} {\bibfnamefont {P.~J.}\ \bibnamefont
  {Holmes}}\ and\ \bibinfo {author} {\bibfnamefont {D.~A.}\ \bibnamefont
  {Rand}},\ }\href@noop {} {\bibfield  {journal} {\bibinfo  {journal} {Journal
  of Sound and Vibration}\ }\textbf {\bibinfo {volume} {44}},\ \bibinfo {pages}
  {237} (\bibinfo {year} {1976})}\BibitemShut {NoStop}%
\bibitem [{\citenamefont {Moon}\ and\ \citenamefont
  {Holmes}(1979)}]{MoonHolmes-1979}%
  \BibitemOpen
  \bibfield  {author} {\bibinfo {author} {\bibfnamefont {F.~C.}\ \bibnamefont
  {Moon}}\ and\ \bibinfo {author} {\bibfnamefont {P.~J.}\ \bibnamefont
  {Holmes}},\ }\href@noop {} {\bibfield  {journal} {\bibinfo  {journal}
  {Journal of Sound and Vibration}\ }\textbf {\bibinfo {volume} {65}},\
  \bibinfo {pages} {275} (\bibinfo {year} {1979})}\BibitemShut {NoStop}%
\bibitem [{\citenamefont {Jordan}\ and\ \citenamefont
  {Smith}(2007)}]{Jordan-book}%
  \BibitemOpen
  \bibfield  {author} {\bibinfo {author} {\bibfnamefont {D.~W.}\ \bibnamefont
  {Jordan}}\ and\ \bibinfo {author} {\bibfnamefont {P.}~\bibnamefont {Smith}},\
  }\href@noop {} {\emph {\bibinfo {title} {Nonlinear ordinary differential
  equations – An introduction for scientists and engineers}}},\ \bibinfo
  {edition} {4th}\ ed.\ (\bibinfo  {publisher} {Oxford University Press},\
  \bibinfo {year} {2007})\BibitemShut {NoStop}%
\bibitem [{\citenamefont {Menard}\ \emph {et~al.}(2000)\citenamefont {Menard},
  \citenamefont {Letellier}, \citenamefont {Maquet}, \citenamefont {Sceller},\
  and\ \citenamefont {Gouesbet}}]{Menard-2000}%
  \BibitemOpen
  \bibfield  {author} {\bibinfo {author} {\bibfnamefont {O.}~\bibnamefont
  {Menard}}, \bibinfo {author} {\bibfnamefont {C.}~\bibnamefont {Letellier}},
  \bibinfo {author} {\bibfnamefont {J.}~\bibnamefont {Maquet}}, \bibinfo
  {author} {\bibfnamefont {L.~L.}\ \bibnamefont {Sceller}}, \ and\ \bibinfo
  {author} {\bibfnamefont {G.}~\bibnamefont {Gouesbet}},\ }\href@noop {}
  {\bibfield  {journal} {\bibinfo  {journal} {International Journal of
  Bifurcation and Chaos}\ }\textbf {\bibinfo {volume} {10}},\ \bibinfo {pages}
  {1759–1772} (\bibinfo {year} {2000})}\BibitemShut {NoStop}%
\bibitem [{\citenamefont {Munkres}(2015)}]{Munkres-2015}%
  \BibitemOpen
  \bibfield  {author} {\bibinfo {author} {\bibfnamefont {J.}~\bibnamefont
  {Munkres}},\ }\href@noop {} {\emph {\bibinfo {title} {Topology}}},\ \bibinfo
  {edition} {2nd}\ ed.\ (\bibinfo  {publisher} {Pearson},\ \bibinfo {year}
  {2015})\BibitemShut {NoStop}%
\bibitem [{\citenamefont {Wasserman}(2024)}]{Wasserman_functionspaces}%
  \BibitemOpen
  \bibfield  {author} {\bibinfo {author} {\bibfnamefont {L.}~\bibnamefont
  {Wasserman}},\ }\href@noop {} {\enquote {\bibinfo {title} {Function
  spaces},}\ } (\bibinfo {year} {2024}),\ \bibinfo {note} {[Downloaded from
  www.stat.cmu.edu/~larry/=sml/functionspaces.pdf on 19 January
  2024]}\BibitemShut {NoStop}%
\bibitem [{\citenamefont {Olver}(2020)}]{Olver-PDE-2020}%
  \BibitemOpen
  \bibfield  {author} {\bibinfo {author} {\bibfnamefont {P.~J.}\ \bibnamefont
  {Olver}},\ }\href@noop {} {\emph {\bibinfo {title} {Introduction to Partial
  Differential Equations}}}\ (\bibinfo  {publisher} {Springer},\ \bibinfo
  {year} {2020})\BibitemShut {NoStop}%
\bibitem [{\citenamefont {Rainville}\ and\ \citenamefont
  {Bedient}(1981)}]{Rainville-1983}%
  \BibitemOpen
  \bibfield  {author} {\bibinfo {author} {\bibfnamefont {E.~D.}\ \bibnamefont
  {Rainville}}\ and\ \bibinfo {author} {\bibfnamefont {P.~E.}\ \bibnamefont
  {Bedient}},\ }\href@noop {} {\emph {\bibinfo {title} {Elementary Differential
  Equations}}},\ \bibinfo {edition} {6th}\ ed.\ (\bibinfo  {publisher}
  {Macmillan Publishing Co.},\ \bibinfo {year} {1981})\BibitemShut {NoStop}%
\bibitem [{\citenamefont {Larochelle}\ \emph {et~al.}(2008)\citenamefont
  {Larochelle}, \citenamefont {Erhan},\ and\ \citenamefont
  {Bengio}}]{Larochelle-2008}%
  \BibitemOpen
  \bibfield  {author} {\bibinfo {author} {\bibfnamefont {H.}~\bibnamefont
  {Larochelle}}, \bibinfo {author} {\bibfnamefont {D.}~\bibnamefont {Erhan}}, \
  and\ \bibinfo {author} {\bibfnamefont {Y.}~\bibnamefont {Bengio}},\ }in\
  \href@noop {} {\emph {\bibinfo {booktitle} {Proc. of the 23rd AAAI Conf. on
  Artif. Intell.}}}\ (\bibinfo {year} {2008})\ pp.\ \bibinfo {pages}
  {646--651}\BibitemShut {NoStop}%
\bibitem [{\citenamefont {Runge}(1895)}]{Runge-1895}%
  \BibitemOpen
  \bibfield  {author} {\bibinfo {author} {\bibfnamefont {C.}~\bibnamefont
  {Runge}},\ }\href@noop {} {\bibfield  {journal} {\bibinfo  {journal} {Math.
  Ann.}\ }\textbf {\bibinfo {volume} {46}},\ \bibinfo {pages} {167} (\bibinfo
  {year} {1895})}\BibitemShut {NoStop}%
\bibitem [{\citenamefont {Kutta}(1901)}]{Kutta-1901}%
  \BibitemOpen
  \bibfield  {author} {\bibinfo {author} {\bibfnamefont {W.}~\bibnamefont
  {Kutta}},\ }\href@noop {} {\bibfield  {journal} {\bibinfo  {journal}
  {Zeitschrift f\"ur Math. u. Phys.}\ }\textbf {\bibinfo {volume} {46}},\
  \bibinfo {pages} {435} (\bibinfo {year} {1901})}\BibitemShut {NoStop}%
\bibitem [{\citenamefont {Heun}(1900)}]{Heun-1900}%
  \BibitemOpen
  \bibfield  {author} {\bibinfo {author} {\bibfnamefont {K.}~\bibnamefont
  {Heun}},\ }\href@noop {} {\bibfield  {journal} {\bibinfo  {journal}
  {Zeitschrift f\"ur Math. u. Phys.}\ }\textbf {\bibinfo {volume} {45}},\
  \bibinfo {pages} {23} (\bibinfo {year} {1900})}\BibitemShut {NoStop}%
\end{thebibliography}%

\end{document}